\documentclass[sigconf]{acmart}
\usepackage{algorithmic, array}
\usepackage[ruled, vlined, commentsnumbered, linesnumbered]{algorithm2e}
\usepackage{mathdots}
\usepackage{amsfonts}
\usepackage{graphicx}
\usepackage{CJKutf8}
\usepackage[normalsize]{subfigure}
\usepackage{wrapfig}
\usepackage{mathdots}
\usepackage{appendix}
\usepackage{multirow}
\usepackage{graphicx}
\usepackage{subfigure}
\usepackage{booktabs}

\newcommand{\DEL}[1]{\iffalse #1 \fi}
\newcommand{\rd}{\color{red}}

\newcommand{\squishlist}{
\begin{list}{$\bullet$}
  { \setlength{\itemsep}{0pt}
     \setlength{\parsep}{0pt}
     \setlength{\topsep}{0pt}
     \setlength{\partopsep}{0pt}
     \setlength{\leftmargin}{0em}
     \setlength{\labelwidth}{0em}
     \setlength{\labelsep}{0.2em} } }

\newcommand{\squishlisttwo}{
\begin{list}{$\bullet$}
  { \setlength{\itemsep}{0pt}
     \setlength{\parsep}{0pt}
    \setlength{\topsep}{0pt}
    \setlength{\partopsep}{0pt}
    \setlength{\leftmargin}{2em}
    \setlength{\labelwidth}{1.5em}
    \setlength{\labelsep}{0.5em} } }

\newcommand{\squishend}{
  \end{list}  }

\newcommand{\YH}[1]{{\color{red}{(Y.H.: #1)}}}
\newcommand{\CQ}[1]{{\color{blue}{(C.Q.: #1)}}}

\newtheorem{definition}{\bf{Definition}}[section]
\AtBeginDocument{%
  }

\copyrightyear{2024}
\acmYear{2024}
\setcopyright{acmcopyright}\acmConference[SIGSPATIAL '24]{The 30th International Conference on Advances in Geographic Information Systems}{October 29 - November 1, 2024}{Atlanta, GA, USA}
\acmBooktitle{The 32nd International Conference on Advances in Geographic Information Systems (SIGSPATIAL '24), October 29 - November 1, 2024}
\acmPrice{15.00}
\acmDOI{10.1145/3557915.3560938}
\acmISBN{978-1-4503-9529-8/22/11}

\begin{document}
\begin{CJK}{UTF8}{gbsn}
  
\title{
Harnessing LLMs for Cross-City OD Flow Prediction  \\
}

\author{Chenyang Yu, Xinpeng Xie, Yan Huang, and Chenxi Qiu}
\email{{chenyangyu, xinpengxie}@my.unt.edu}
\email{{yan.huang, chenxi.qiu}@unt.edu} 
\affiliation{%
  \institution{Department of Computer Science and Engineering, University of North Texas}
  \streetaddress{P.O. Box 1212}
  \city{Denton}
  \state{Texas}
  \country{USA}
  \postcode{43017-6221}
}

\renewcommand{\shortauthors}{Chenyang Yu, Xinpeng Xie, Yan Huang, and Chenxi Qiu}

\begin{abstract}
Understanding and predicting \emph{Origin-Destination (OD)} flows is crucial for urban planning and transportation management. 
Traditional OD prediction models, while effective within single cities, often face limitations when applied across different cities due to varied traffic conditions, urban layouts, and socio-economic factors. \looseness = -1

In this paper, by employing \emph{Large Language Models (LLMs)}, we introduce a new method for cross-city OD flow prediction. Our approach leverages the advanced semantic understanding and contextual learning capabilities of LLMs to bridge the gap between cities with different characteristics, providing a robust and adaptable solution for accurate OD flow prediction that can be transferred from one city to another. Our novel framework involves four major components: collecting OD training datasets from a source city, instruction-tuning the LLMs, predicting destination POIs in a target city, and identifying the locations that best match the predicted destination POIs. We introduce a new loss function that integrates POI semantics and trip distance during training.  By extracting high-quality semantic features from human mobility and POI data, the model understands spatial and functional relationships within urban spaces and captures interactions between individuals and various POIs. 
Extensive experimental results demonstrate the superiority of our approach over the state-of-the-art learning-based methods in cross-city OD flow prediction. 

\end{abstract}

\maketitle

\section{Introduction}

Origin and destination (OD) matrices are invaluable tools in transportation planning and urban development. By capturing the flow of people, goods, or services between different locations within a geographic area, OD matrices provide critical insights into travel patterns, demand for infrastructure, and spatial interactions. Urban planners and policymakers utilize OD matrices to identify transportation bottlenecks \cite{cerqueira2022inference, sun2021demand}, optimize transit routes \cite{afandizadeh2021hourly, gentile2016modelling}, and allocate resources efficiently \cite{rong2023complexity, sun2021demand}. 

Gathering accurate and comprehensive data on travel movements within a geographic area can be time-consuming and costly. Data collection efforts may face challenges such as incomplete or inconsistent data, limited data availability for certain modes of transportation or population groups, and privacy concerns related to personal travel data.

Traditional OD estimation models such as \cite{lenormand2016systematic} and the Radiation Model \cite{ren2014predicting} rely on the assumptions that flow of people or goods between two locations is proportional to their populations and inversely proportional to the square of their distance. They have been adapted to incorporate novel data sources such as social media check-ins \cite{pourebrahim2019trip}, enhancing their adaptability. 

Recent learning-based models excel in capturing implicit data characteristics and delivering commendable performance\cite{9657493, 9805695} \DEL{\YH{These should be consistent with the papers we referred in related work. Please update}}. While effective in achieving their objectives, above works primarily focus on the OD flow generation within a single city. Some recent works  \cite{rong2023goddag} propose to generate trajectories across cities by learning traffic patterns and vehicle behaviors from data-rich cities and extrapolating this information to generate datasets for cities lacking such data. However, it heavily depends on the vehicle trajectory and road network datasets specific to the target city. Transferring the learned patterns from these datasets to different cities is challenging due to variations in traffic conditions and urban layouts among cities.

Large language models (LLMs) \cite{brown2020language, Devlin2019BERTPO, radford2019language, han2024chainofinteractionenhancinglargelanguage, radford2018improving} have demonstrated impressive capabilities across diverse tasks by recognizing patterns in vast datasets, enabling them to achieve high accuracy in various applications, from natural language processing to fields such as biomedical imaging and crop management \cite{Lai_2024_CVPR, Wu_2024_CVPR}. An interesting question is whether LLMs can understand and predict likely trip destinations based on a given point of interest and timeframe in the same way humans do. Has some base knowlege about trip behaviors learned across cities and once fine-tuned for flow prediction in one city and can be easily transfered to a different city? These capabilities hinge on the LLMs' ability to consider various factors and make informed guesses about trip destination as well as recognize patterns and associations in trip data. The main challenges of leveraging an LLM model are that: (1) the models were pre-trained on texts and do not have explict knowlege of OD flows; (2) The primary objective of the loss function of LLMs is to improve the model's ability to predict the next token in a sequence accurately, thereby enhancing its performance in generating coherent and contextually relevant text. We need to design new loss functions that take the spatial context into consideration. 

\subsection{Our Contributions}
In this paper, we propose a framework to investigate and harness large language models' capabilities in understanding trips for OD flow prediction across cities.
Our approach is driven by the following premises: 
(1) despite the considerable variations in their trips across different cities,  
people travel for specific purposes and tend to follow similar patterns in their Origin-Destination POIs \cite{lee2015relating, huang2018mining} and (2) LLMs are trained on vast amounts of text data, including information about transportation, urban planning, and human behavior, and has basic understanding of city mobility patterns.

\begin{itemize}

\item{\bf Contribution 1}. We propse an \emph{\underline{LLM} based \underline{C}ross-city \underline{OD} (LLM-COD)} flow prediction framework. LLM-COD leverages the knowledge possed by LLMs and enhance it to infuse understanding of  origin and destination related to point of interests, time, and distance. 
LLM-COD captures how people interact with various POIs in the city related to time.

\item{\bf Contribution 2}. We develop a process to allow
LLM-COD to transfer the knowledge to other cities and perform OD flow prediction with adaptation. We propose a new loss function that integrates POI semantics with distance which helps LLM-COD to achieve reboust performance across cities. This means that even with relatively limited training data for a new city, the model can utilize the knowledge and mobility patterns learned from other cities to improve prediction accuracy.

\item{\bf Contribution 3}. We conducted extensive experiments to evaluate our method. The results demonstrate that our approach significantly outperforms traditional models and state-of-the-art learning-based methods in cross-city OD flow prediction. LLM-COD reduces RMSE significantly compared to the best state-of-the-art model espcially for challenging high precision OD flow  prediction (reduces by ~46\% for 1,000m $\times$ 1,000m) and is consistently better in all other commonly used metrics. 
\end{itemize}

\section{Overview}
We consider a scenario where one city, referred to as "City $A$", possesses an Origin-Destination (OD) dataset, which includes the origin and destination points of various vehicle trips. In contrast, the other city, referred to as "City $B$", does not have access to the OD dataset. Our objective is to utilize the OD dataset from City $A$ to train a generative model that can subsequently produce an OD dataset for City $B$.

Our approach is motivated by the observation that, despite significant variations in individuals' trips across different urban environments---attributable to varied transportation modes and city layouts---\textit{people in different cities exhibit similar patterns in their Origin-Destination Points of Interest (POIs)}. For example, a universal OD pattern is the daily commute between residential areas and workplaces \cite{Kevin-PLOSONE2014}. Additionally, educational institutions like schools, colleges, and universities serve as major OD points in cities, creating distinct travel patterns tied to the academic year, with weekdays seeing increased travel during morning and afternoon peaks.

In recent years, \textit{deep neural networks (DNNs)} have proven to be highly effective in modeling and analyzing daily behavior patterns, owing to their ability to discern complex patterns from vast datasets \cite{ISLAM2022306}. The advancement of LLMs has notably transformed many fields and enabled  deeper insights into human behavior and trends. An LLM is initially pre-trained on extensive data collections and subsequently adapted to address specific downstream tasks. While initial applications were focused on tasks such as language translation \cite{NIPS2017_3f5ee243} and sentiment analysis \cite{Devlin2019BERTPO}, the scope of LLMs has broadened to include a variety of fields, including the prediction of human behavior \cite{kang2023values}.

\begin{figure}[t]
\centering
\hspace{0.00in}
\begin{minipage}{0.49\textwidth}
  \subfigure{
\includegraphics[width=1.00\textwidth]{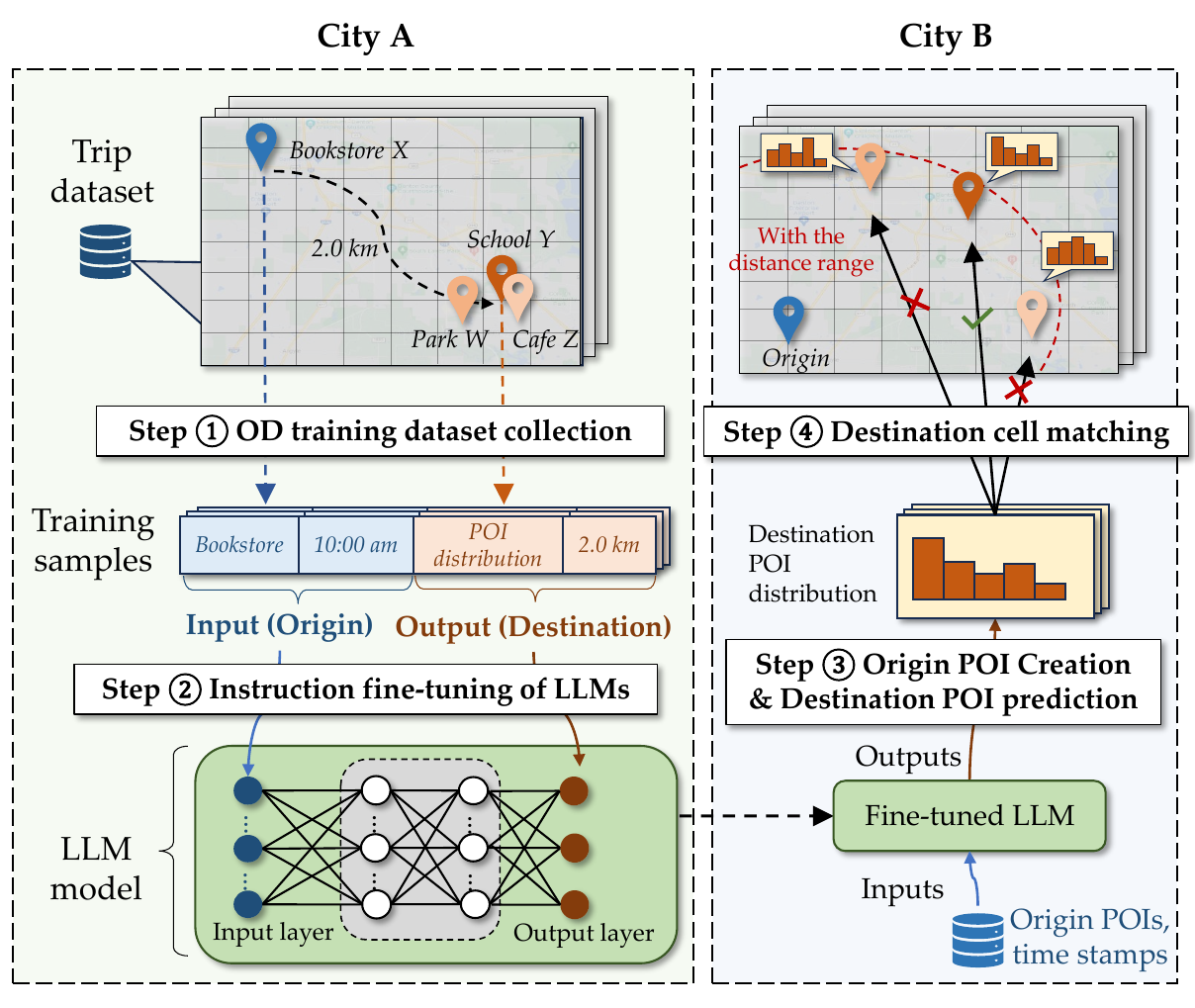}}
\vspace{-0.00in}
\end{minipage}
\caption{LLM-COD Framework.}
\label{fig:framework}
\vspace{-0.00in}
\end{figure}

Leveraging the strong capability of LLMs to learn human behavior patterns, LLM-COD is designed to generate OD datasets across various cities. Figure \ref{fig:framework} shows the framework of our approach, composed of the following four steps \textcircled{1}--\textcircled{4}:  
\begin{itemize}
\item [\textcircled{1}] \textbf{OD POI training dataset collection (in City $A$):} Given the OD vehicle trip dataset from City $A$, we construct an OD POI training dataset. Each sample in this dataset consists of inputs, which include a list of origin POIs and the associated timestamp, and outputs, which comprise a list of destination POIs and the permissible distance range between the origin and destination. 
\item [\textcircled{2}] \textbf{Instruction fine-tuning of LLM}: We take a pre-trained LLM (Llama2 + Gemma) and fine-tune the model using the OD training dataset created in \textcircled{1}. \textcircled{2} leverages the general understanding and linguistic capabilities the LLM model has learned from a vast and diverse initial training phase and then narrows its focus to our OD POI generation problem.
\item [\textcircled{3}] \textbf{Origin POI creation and their destination POI prediction (in City $B$)}: Assuming City $B$ possesses the origin POIs along with their time stamps, the generative LLM model fine-tuned in \textcircled{2} is capable of predicting a distribution of destination POIs for each specified origin POI, as well as the acceptable distance range. 
\item [\textcircled{4}] \textbf{Destination cell matching (in City $B$)}: The resulting lists of destination POIs can then be aligned with the most similar location (represented as a grid cell), based on the highest similarity in POI distributions.
\end{itemize}



\DEL{
Our approach is driven by the observation that, despite the considerable variations in people's trajectories across different cities - attributable to diverse transportation modes and urban layouts - \textbf{people in different cities tend to follow similar patterns in their Origin-Destination POIs}. For instance ({\rd Chenxi: the following examples could be shortened}): 
\newline \textbf{Home-to-work commutes}: The need to travel between residence and workplace is universal (one of the most consistent OD patterns across cities).
\newline \textbf{Educational facilities}: Schools, colleges, and universities are significant OD points in urban areas, leading to specific travel patterns related to the academic calendar, with weekdays experiencing higher travel volumes during morning and afternoon times.
\newline \textbf{Central business districts (CBDs)}: Many cities have a central business district or downtown area that serves as a major destination for workers, shoppers, and tourists. This concentration of economic activities creates a common pattern of people traveling towards these areas, especially during morning and evening rush hours.
\newline \textbf{Shopping and errands}: Retail centers, including shopping malls, markets, and commercial strips, attract a significant number of visitors, contributing to predictable OD flows, particularly on weekends and during special shopping seasons.
\newline \textbf{Recreational and leisure activities}: Locations such as parks, cultural centers, sports facilities, and entertainment venues generate distinct travel patterns, often characterized by higher weekend and evening traffic.
\newline \textbf{Public Transportation Nodes}: Cities with developed public transportation systems show pronounced OD patterns centered around major transit hubs, including train stations, bus terminals, and metro stations, facilitating the movement of large numbers of people.
\newline \textbf{Temporal Patterns}: Similarities in daily (peak hours in the morning and evening), weekly (higher activity on weekdays for work and school, weekends for leisure), and seasonal patterns (influenced by school terms, holidays, and weather conditions) are observed across cities.}


\section{Methodology}

In this section, we detail the four steps for generating the OD dataset of City $B$ based on City $A$ (\textbf{Section \ref{subsec:ODtraining}--Section \ref{subsec:ODmatching}}), as outlined in Figure \ref{fig:framework}. Before that, we first introduce the notations and the definitions in \textbf{Section \ref{subsec:notations}}. Table \ref{Tb:Notationmodel} lists the main notations used throughout this paper.

\subsection{Notations and Definitions}
\label{subsec:notations}
We use the superscript $^X$ to denote the notations specific to City $X$, where $X$ can be either $A$ or $B$.
\begin{definition} 
\label{def:Grid}
(Grid) City $X$ is divided into $N^X$ square cells of equal size (e.g., 500m × 500m), denoted as $\mathcal{V}^X = \left\{v^X_1, \ldots, v^X_{N^X}\right\}$, where each $v^X_i$ represents the $i$-th grid cell in City $X$. The center coordinates of each grid cell are assumed to be known for both cities, as further discussed in Section \ref{subsec:ODtraining}.
\end{definition}

\begin{definition}
\label{def:Spatial Features}
(Spatial features) We consider $K$ distinct types of POIs, detailed in Table \ref{tab:poi_frequency} in the Appendix. Each grid cell $v^X_i$ ($i = 1, ..., N^X$) can be characterized by a spatial feature vector $\mathbf{u}^X_i  = \left[u^X_{i,1}, \ldots, u^X_{i,K}\right]$, where each $u^X_{i,k}$ ($k = 1, ..., K$) denotes the number of the $k$-th type of POIs in grid cell $v^X_i$. We use $\mathcal{U}^X = \left\{u^X_1, ..., u^X_{N^X}\right\}$ to represent the spatial features of the grid cells of City X. 

\DEL{\CQ{Chenyang, Xinpeng, add here [for instance, the POIs of cities can be retrieved by xxx, insert citation here] addressed}.} 
\end{definition}

\begin{definition}
\label{def: trip set}
(Trip Set) We represent each trip $l$ by a 4-tuple $\left(v^X_{O_l}, t^X_l, v^X_{D_l}, c^X_l\right)$, including its origin cell $v^X_{O_l}$, the starting time $t^X_l$, its destination cell $v^X_{D_l}$, and the acceptable traveling cost $c^X_l$. We use $\mathcal{T}^X = \left\{\left(v^X_{O_l}, t_l, v^X_{D_l}, c^X_l\right)\right\}_{l = 1, ...,L^X}$ to represent the trip set of City $X$, where $L^X$ is the number of trips in $\mathcal{T}^X$. 
\end{definition} 

\begin{definition}
\label{def: origin set}
(Origin Set) The origin set of city $X$, denoted as $\mathcal{O}^X$, is defined as the set of origin cells, along with their corresponding starting time points for the trips $\mathcal{T}^X$ in City $X$, i.e., $\mathcal{O}^X = \left\{\left(v^X_{O_l}, t^X_l\right)\right\}_{l = 1, \ldots, L^X}$.
\end{definition}

\begin{definition}
\label{def:OD pair}
(OD pair learning) Given an origin set $\mathcal{O}^X$, OD pair learning is a process of predicting the destination $\hat{v}^X_{D_l}$ and traveling cost $\hat{c}^X_{l}$ of each $\left(v^X_{O_l}, t^X_l\right) \in \mathcal{O}^X$, creating the corresponding trip $\left(v^X_{O_l}, t^X_l, \hat{v}^X_{D_l}, \hat{c}^X_l\right)$. 
\end{definition}

\DEL{
\begin{definition}
\label{def: trip set}
(Trip Set) Trip set of city $X$ is a set of trips given only the origin cells and starting time $\mathcal{T}^X = \left\{\left(v^X_{O_l}, t^X_l\right)\right\}$
\end{definition}}

\begin{definition}
\label{def:OD flow}
(OD flow) The OD flow is the volume of movement between two grid cells. We use $\mathcal{T}^X_{ij}$ to represent the set of trips from grid cell $v^X_i$ to $v^X_j$
\begin{equation}
    \mathcal{T}^X_{ij} = \left\{\left(v^X_{O_l}, t^X_l, v^X_{D_l}, c^X_l\right)\in \mathcal{T}^X \mid v^X_{O_l} = v^X_i, v^X_{D_l} = v^X_j \right\}, 
\end{equation}
and use $f^X_{ij} = \left|\mathcal{T}^X_{ij}\right|$ to represent the OD flow from cell $v^X_i$ to $v^X_j$. 
\end{definition}

\begin{definition}
\label{def:OD matrix}
(OD matrix) The OD matrix $\mathbf{F}^X = \left\{f^X_{ij} \right\}_{N^X\times N^X}$ includes
the OD flows between each pair of grid cells in City X.
\end{definition}

\paragraph{Problem formulation.} The cross-city OD generation problem can be considered as a domain adaptation problem\cite{ben2010theory}, which aims at learning a model from a source data distribution and transferring the knowledge to a different (but related) target data distribution. Given the above notations and definitions (Definition \ref{def:Grid} -- \ref{def:OD matrix}), We formally define the cross-city OD generation problem in \emph{Definition \ref{def:Problem}}:
\begin{definition}
\label{def:Problem}
(Cross-city OD generation problem) 
\begin{itemize}
\item \textbf{Instance}: Given the location sets of both cities, $\mathcal{V}^A$, $\mathcal{V}^B$, their spatial features $\mathcal{U}^A, \mathcal{U}^B$, the trip set of City $A$, $\mathcal{T}^A$, and the origin set of City $B$, $\mathcal{O}^B$; 
\item \textbf{Question}: How to predict the OD flow matrix $\hat{\mathbf{F}}^B$ of City $B$.
\end{itemize}
\end{definition}
Note that given the learned OD pairs (Definition \ref{def:OD pair}) at City $B$, we can count the number of trips between each pairs of cells in City $B$, therefore obtaining the OD flow matrix (according to Defintion \ref{def:OD flow} and Defintion \ref{def:OD matrix}). Next, we detail our approach, LLM-COD, for the OD pair learning, including the four steps outlined in Figure \ref{fig:framework}.

\begin{table}[t]
\caption{Main notations and their descriptions}
\vspace{-0.00in}
\label{Tb:Notationmodel}
\centering
\normalsize 
\small 
\begin{tabular}{l l}
\specialrule{1.0pt}{0pt}{0pt} 
Symbol                  & Description \\
\hline
\hline
$\mathcal{V}^X$          & The set of grid cells of City X\\
$\mathcal{T}^X$ & The trip set in City X \\
$N^X$ & The total number of discrete locations in $\mathcal{V}^X$ \\
$\mathcal{O}^X$ & The origin set in City X \\ 
$K$ & The total number of POI types \\ 
$L^X$ & The total number of trips in City $X$\\
$v^X_i$           & The $i$th grid cell in $\mathcal{V}^X$ \\
$\sigma^X$           & The grid cell size of City $X$ \\
$\mathbf{u}^X_i$           & $\mathbf{u}^X_i = \left[u^X_{i,k}, ..., u^X_{i,K}\right]$ denotes the spatial feature of the \\
& cell $v^X_i$, where each $u^X_{i,k}$ denotes the number of type $k$ POIs \\
& within $v^X_i$.  \\ 
$\hat{\mathbf{u}}^X_i$           & The predicted spatial feature of the cell $v^X_i$ \\
$v^X_{O_l}$ & The origin cell of trip $l$ \\
$v^X_{D_l}$ & The destination cell of trip $l$ \\
$c^X_l$ & The travel cost of trip $l$ \\
$t^X_l$ & The starting time of trip $l$ \\ 
$f(\cdot)$ & The function describing the relationship between origin \\
& and destination cells in each trip \\
\hline
\end{tabular}
\normalsize
\vspace{-0.00in}
\end{table}

\subsection{Step \textcircled{1} - OD Training Dataset Collection in City $A$}
\label{subsec:ODtraining}
To construct an \emph{origin-destination POI} dataset for both Cities $A$ and $B$, we follow a streamlined process encompassing the following three steps, as illustrated in Figure \ref{fig:data_retrieval}: 
\begin{figure}[t]
\centering
\hspace{0.00in}
\begin{minipage}{0.50\textwidth}
  \subfigure{
\includegraphics[width=1.00\textwidth]{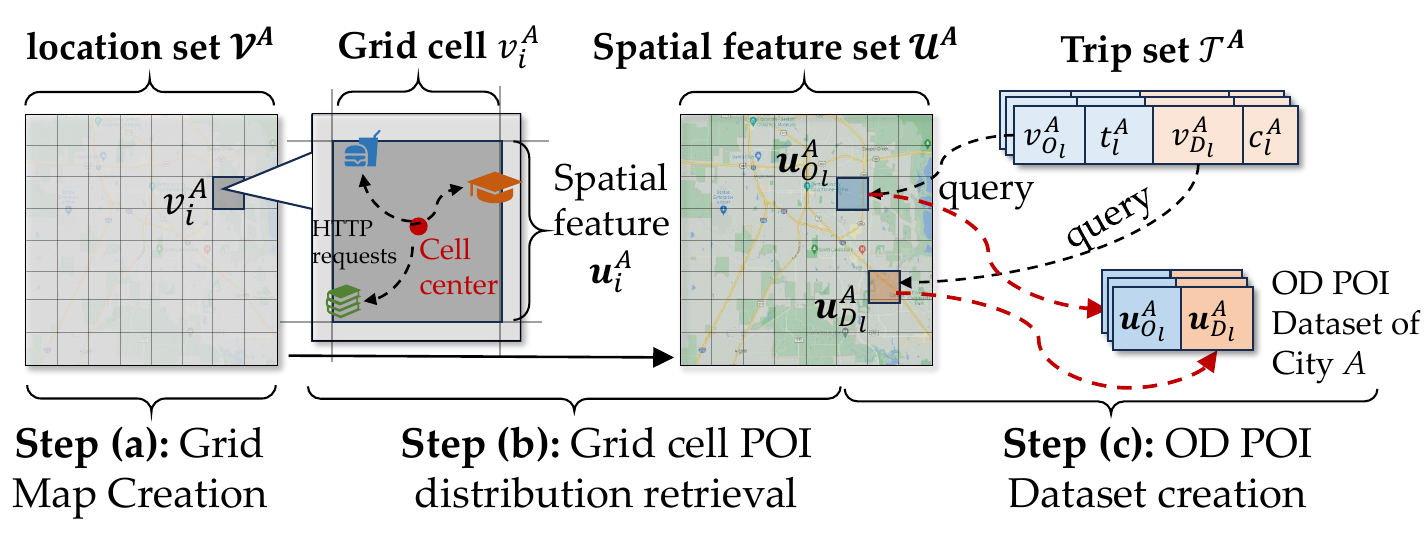}}
\vspace{-0.15in}
\end{minipage}
\caption{OD training dataset collection.}
\label{fig:data_retrieval}
\vspace{-0.1in}
\end{figure}
\begin{enumerate}
    \item [(a)] \textbf{Grid Map Creation:} Given the geographical boundary coordinates (latitude and longitude) of both Cities $A$ and $B$, we define a rectangular bounding box that encompasses the area of interest. As Figure \ref{fig:data_retrieval} (\textbf{Step (a)}) shows, we  divide this area into square grid cells with size $\sigma^A \times \sigma^A$ (e.g., $\sigma^A =$ 500m, 1000m, and 2000m in our experiments). For simplicity, the cell sizes are kept consistent for both City $A$ and City $B$, i.e., $\sigma^A = \sigma^B$. In this step, we can obtain the location sets of both cities, $\mathcal{V}^A$ and $\mathcal{V}^B$. 
    \item [(b)] \textbf{Grid cell POI distribution retrieval:} We retrieve the POI distributions for each cell to obtain the spatial feature sets $\mathcal{U}^A$ and $\mathcal{U}^B$ of the two cities. As Figure \ref{fig:data_retrieval} (\textbf{Step (b)}) shows, for each grid cell $v^A_i$, we create its spatial feature $\mathbf{u}^A_i$ by identifying the POIs whose nearest center point corresponds to the center point of $v^A_i$. Specifically, in the experiment in Section \ref{sec:performance}, we use asynchronous HTTP requests to query nearby POIs for each cell's central point via the Map Service API. 
    
    \item [(c)] \textbf{OD POI dataset creation:} We then process the trajectory dataset to obtain the POI distributions of the origin and destination cells of each trip. As Figure \ref{fig:data_retrieval} (\textbf{Step (c)}) shows, for each trip $l$ in City $A$, we query POI types for its OD cells $\left(v^A_{O_l}, v^A_{D_l}\right)$ from $\mathcal{U}^A$ created in \textbf{Step (b)}, obtaining their corresponding POI distributions $\mathbf{u}^A_{O_l}$ and $\mathbf{u}^A_{D_l}$, respectively. The whole set of OD POI distributions $\left\{\left(\mathbf{u}^A_{O_l}, \mathbf{u}^A_{D_l}\right)\right\}_{l = 1, ..., L^A}$ is saved for further analysis and also served as the training dataset for fine-tuning the LLM models. 
\end{enumerate}

\subsection{Step \textcircled{2} - Instruction-Tuning of LLMs}
\label{subsec:finetune}

Our next step is to fine-tune an LLM using instructions to establish the OD pair learning in \emph{textual format}.


However, most existing DNN-based approaches for predicting traffic flows still rely primarily on numerical data, such as road network and node features \cite{rong2023goddag, DapengZhang}. As a solution, we guide the LLM with a specific instruction, as detailed in Table \ref{table:instruction}, which specifies the spatial features of the origin and destination, thereby enhancing the LLM's understanding the input format, and what the responses (or the outputs) are expected.


\begin{table}[h!]
\caption{Instruction format}
\label{table:instruction}
    \centering
    \small 
    \begin{tabular}{p{1.5cm}|p{6.5cm}}
        \specialrule{1.0pt}{0pt}{0pt} 
        \textbf{Instruction} & 
        \textcolor{red}{<Instruction> \emph{Given the starting place and time of a taxi trajectory in [city], predict the most likely destination and how far it is from the starting point.}}
        
        \textcolor{red}{\emph{Please use the provided "Candidate POIs" list to describe the starting place and destination.}}
        
        \textcolor{red}{\emph{Candidate POIs: [POI list]}} 
        \\ \hline \hline
        \textbf{Input \newline Sample} & 
        \textcolor{orange}{<Input> Starting place: [Food \& Cuisine, Healthcare, Shopping, Tourist Attraction], Starting time: [12:35]}
         \\ \hline
        \textbf{Response Sample} & \textcolor{blue}{"POIs": [Residential area, Company, Hotel, Cultural Venue, Shopping],
            "traveling cost": [1.3 kilometers]
        } \\ \hline

        \textbf{Predicted Cell Index} & {23}
        \\ \hline
        
    \end{tabular}
\end{table}

\begin{figure}[t]
\centering
\hspace{0.00in}
\begin{minipage}{0.48\textwidth}
  \subfigure{
\includegraphics[width=1.00\textwidth]{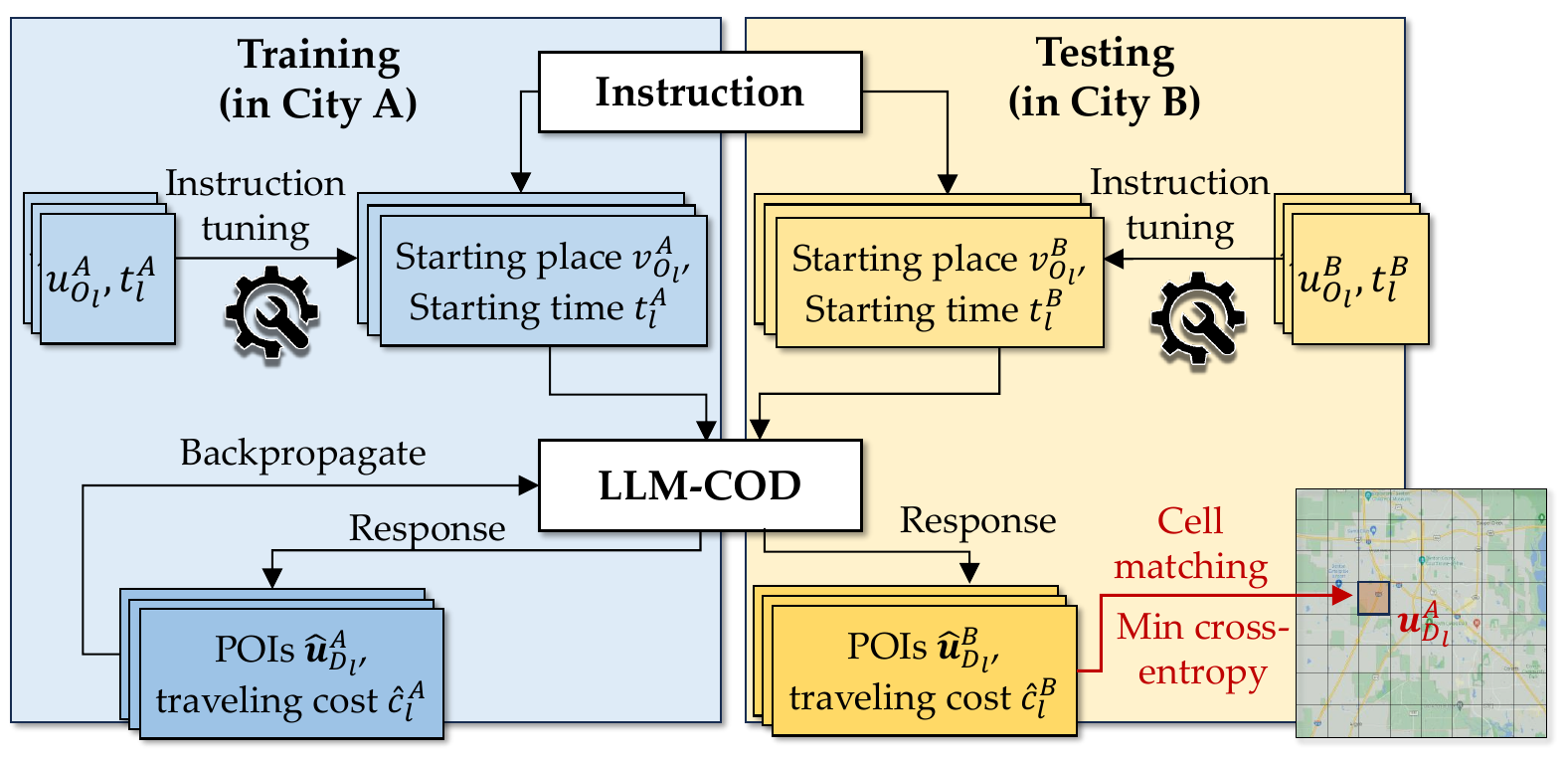}}
\vspace{-0.15in}
\end{minipage}
\caption{LLM-COD.}
\label{fig:LLM}
\vspace{-0.10in}
\end{figure}

As depicted in Figure \ref{fig:LLM}, we apply the LLM  to a cross city OD flow prediction scenario
guided by the instruction format in Table \ref{table:instruction}. The primary functions of LLM-COD are twofold:
\begin{enumerate}
\item[(a)] \textbf{Constructing input prompt}. Specifically,
for each OD trip, given $\left(u_{O_l}^A, t_l^A\right)$, we form the input sample as "<Input> Starting place: $\left[v_{O_l}^A\right]$, Starting time: $\left[t_l^X\right]$". 
\item[(b)] \textbf{Training and inferencing}. For training, after we get "POIs" and "traveling cost", we will combine the two objects together and backpropagate the loss to update the parameters. For inference, we will perform the destination cell matching using the two objects, introduced in step \textcircled{4}.
\end{enumerate}

We use a function $f$ to represent the relationship between each OD pair in the LLM model, 
\begin{equation}
\label{eq:fLLM}
\left(\hat{\mathbf{u}}^X_{D_l}, \hat{c}^X_l\right) = f\left(\mathbf{u}^X_{O_l}, t^X_l; \boldsymbol\theta\right). 
\end{equation}
where 
\begin{itemize}
    \item the \textbf{inputs} include the spatial feature of the origin cell $\mathbf{u}^X_{O_l}$ and the starting time stamp $t^X_l$, and 
    \item the \textbf{outputs} include the spatial feature of the destination cell $\hat{\mathbf{u}}^X_{D_l}$ and the accepted traveling cost $\hat{c}^X_l$ between the origin cell and destination cell:
\end{itemize} 
Here, $\boldsymbol\theta$ is the parameters of $f$. Our goal is to optimize $\boldsymbol\theta$, achieved by fine-tuning the Language model. 

\textbf{Low-rank adaptation}. We apply \emph{Low-Rank Adaptation (LoRA)} \cite{hu2021lora}, freezing dense layers
in LLMs and updating weights with rank decomposition matrices. In LoRA, a hidden layer weight update transformation could be represented as 
\begin{equation}
h = \mathbf{W}_0 x + \Delta \mathbf{W} x = \mathbf{W}_0 x + \mathbf{B}\mathbf{A} x, 
\end{equation}
where $\mathbf{W}_0 \in \mathbb{R}^{d \times k}$ is the original weight matrix, and $\mathbf{B} \in \mathbb{R}^{d \times r}$ and $\mathbf{A} \in \mathbb{R}^{r \times k}$ are the low-rank matrices with $r \ll \min(d, k)$. This low-rank adaptation effectively captures the necessary adjustments without the overhead of updating the entire model, as seen in its application to geoscience and other fields. 

\textbf{Loss function}. During training, we aim to predict the spatial feature of each destination cell within the target urban area and the acceptable traveling cost to reach the destination cell. As shown in Figure \ref{fig:LLM}, LLM will output a POI list and a traveling cost. We combine the two objects together and employ \emph{weighted cross-entropy loss} between the actual and predicted outcomes as the loss function, defined by:
\DEL{
\begin{equation}    \text{CrossEntropy}\left(\overline{\hat{\mathbf{u}}}^X_{D_l}, \overline{\mathbf{u}}^X_i\right) = -\sum_{k=1}^{K} w_{k} \overline{u}^X_{D_l, k} \log\left(\overline{\hat{u}}^X_{D_l, k}\right)
\end{equation}
where $L$ denotes the total number of trajectory samples. 
Here, 
$$\overline{\mathbf{u}}^X_{D_l} = \left[\overline{u}^X_{D_l,1}, ..., \overline{u}^X_{D_l,K}\right]~\mbox{and}~\overline{\hat{\mathbf{u}}}^X_{D_l}= \left[\overline{\hat{u}}^X_{D_l,1}, ..., \overline{\hat{u}}^X_{D_l,K}\right]$$ 
represent the normalized true POI distribution and predicted POI distribution for cell $v^A_{D_l}$, respectively. Specifically, each $\overline{u}^A_{D_l,k}$ and $\overline{\hat{u}}^X_{D_l,k}$ are normalized by the soft max function: 
\begin{equation}
\overline{u}^X_{D_l,k} = \frac{e^{u^X_{D_l,k}}}{\sum_{n=1}^{K}e^{u^X_{D_l,n}}},~ \overline{\hat{u}}^X_{D_l,k} = \frac{e^{\hat{u}_{D_l,k}}}{\sum_{n=1}^{K}e^{\hat{u}_{D_l,n}}}
\end{equation}
where $u^X_{D_l,k}$ and $\hat{u}^X_{D_l, k}$ denote the actual and forecasted probabilities of POI $k$ within cell $c_{D_l}$, respectively.


Consider adding distance as an additional entry of the probability distribution in the cross entropy loss function:}
\begin{equation}
\label{eq:crossentropy}
    \text{CrossEntropy} = - \left(\sum_{k=1}^{K} w_{k}\overline{u}^X_{D_l, k} \log(w_{k} \overline{\hat{u}}^X_{D_l, k}) + \alpha \overline{c}^X_{l} \log(\alpha \overline{\hat{c}}^X_{l})\right).
\end{equation}
Here, we use $u^X_{D_l,k}$ and $\hat{u}^X_{D_l, k}$ ($c^X_{l}$ and $\hat{c}^X_{l}$) to denote the actual and forecasted numbers of POI $k$ (actual and forecasted traveling cost) within cell $v^X_{D_l}$, respectively. In Equ. (\ref{eq:crossentropy}), each $u^X_{D_l,k}$, $\hat{u}^X_{D_l,k}$, $c^X_{l}$, and $\hat{c}^X_{l}$ are normalized by the soft max function: 
\begin{equation}
\overline{u}^X_{D_l,k} = \frac{e^{u^X_{D_l,k}}}{\sum_{k=1}^{K}e^{u^X_{D_l,k}} +  e^{c^X_l}}, \overline{c}^X_{l} = \frac{e^{c^X_{l}}}{\sum_{k=1}^{K}e^{u^X_{D_l,k}} + e^{c^X_l}},
\end{equation}
\begin{equation}
\overline{\hat{u}}^X_{D_l,k} = \frac{e^{\hat{u}_{D_l,k}}}{\sum_{k=1}^{K}e^{\hat{u}_{D_l,k}}+ e^{\hat{c}^X_l}}, \overline{\hat{c}}^X_{l} = \frac{e^{\hat{c}^X_{l}}}{\sum_{k=1}^{K}e^{\hat{u}_{D_l,k}} + e^{\hat{c}^X_l}}.
\end{equation}
Each $w_k$ ($k = 1, ..., K$) is the weight assigned to the probability $\overline{u}^X_{D_l, k}$ and $\overline{\hat{u}}^X_{D_l, k}$ and $\alpha$ is the weight assigned to the travel cost $\overline{c}^X_{l}$ and $\overline{\hat{c}}^X_{l}$. 
\DEL{
To analyze the relationship between $\alpha$ and the sensitivity of cross entropy to the error $\delta = |\overline{c}^X_{l} - \overline{\hat{c}}^X_{l}|$, i.e., or check how 
\begin{equation}
\frac{d CrossEntropy(\delta)}{d \delta} 
\end{equation}
is impacted by $\alpha$. It might be hard to find the closed form while it is possible to give numerical results.}

Next, we use the fine-tuned LLM model to generate a set of OD pairs for City $B$ in Step \textcircled{3} and Step \textcircled{4}.

\subsection{Step \textcircled{3} - Origin POI Creation and Their Destination POI Prediction in City $B$}
\label{subsec:ODPOIpredict}
In this step, we first create an origin cell set $\mathcal{O}_B$ in City $B$, incorporating POIs within each cell. $\mathcal{O}_B$ can be established using datasets that are mostly available in major cities, such as (1) demographic data, which reflects the density of origin locations across different areas \cite{Wu-CEUS2019}, (2) requested location data in LBS, which often serve as the starting points for trajectory analysis \cite{QIAN2021101552}, (3) location data from large events, stadiums \cite{Muchisky-PP2002}, or venues, which includes information on where attendees originate, and (4) public transportation data \cite{Yan-Infocom2017}. 

As an example, in the experiment part (Section \ref{sec:performance}), we use Tencent Location Requested (TLR) data between March 1 and March 28, 2024, sourced from Tencent’s big data portal \href{https://heat.qq.com}{(https://heat.qq.com)}. This data captures the number of location requests from Tencent's LBSs, which include services related to social networks, games, online shopping, communication, and travel.  Liu
et al. \cite{QIAN2021101552} found the data exhibits a high correlation coefficient of 0.9 with the residential population, indicating a strong relationship between Tencent’s location data and human activity patterns. This strong correlation suggests that Tencent's location-aware dataset can serve as a proxy for human activities on a short-time scale, making it suitable for defining origin cells in any city in China.

Given each generated origin POI distribution $\mathbf{u}^B_{O_l}$ in City $B$ and their corresponding timestamps $t^B_l$ as the inputs, the fine-tuned LLM (described by Equ. (\ref{eq:fLLM})) predicts the destination POI distribution $\hat{\mathbf{u}}^B_{D_l}$ along with an acceptable distance range $\hat{c}^B_l$ between the origin and destination cell.

\subsection{Step \textcircled{4} - Destination Cell Matching and OD Matrix Calculation in City $B$}
\label{subsec:ODmatching}
For every trip $\left(v^B_{O_l}, t^B_l\right) \in \mathcal{T}^B$ in City $B$, we can utilize LLM to predict the POI distribution $\hat{\mathbf{u}}^B_{D_l}$ of the destination cell and the acceptable traveling cost $\hat{c}^B_l$. To find the matched destination cell, we check each $v^B_i \in \mathcal{V}^B$ where the travel cost from $v^B_{O_l}$ to $v^B_i$ is no higher than $\hat{c}^B_l$. 

Among these cells, we select the one whose normalized POI distribution $\overline{\mathbf{u}}^B_i$ exhibits the lowest cross entropy with the normalized predicted POI distribution $\overline{\hat{\mathbf{u}}}^B_{D_l}$. 
\begin{equation}
\hat{v}_{D_l} = \arg\min_{v^B_i \in \mathcal{V}^B, c(v^B_{O_l}, v^B_i)\leq \hat{c}^B_{l}} \text{CrossEntropy}\left(\overline{\hat{\mathbf{u}}}^X_{D_l}, \overline{\mathbf{u}}^X_i\right)
\end{equation}
Consequently, the set of OD pairs $\left\{\left(v^B_{O_l}, \hat{v}^B_{D_l}\right) \mid v^B_{O_l} \in \mathcal{O}_B .\right\}$ constitutes the OD dataset generated in City $B$.

\section{Performance Evaluation}
\label{sec:performance}
In this section, we conduct extensive experiments to test the performance of our approach, LLM-COD, with the emphasis on answering the following 4 questions:  
\begin{itemize}
\item[RQ1:] What is the performance of OD flow prediction of LLM-COD in new cities compared to the state of arts (\textbf{Section \ref{subsec:RQ1}})? 
\item[RQ2:] How do the design choices of our method impact performance (\textbf{Section \ref{subsec:RQ2}})?  
\item[RQ3:] Can the LLM-COD model robustly handle the predicting tasks with varying OD flow and trip distance distribution (\textbf{Section \ref{subsec:RQ3}})?  
\item[RQ4:] Are the generated OD flows spatially distributed in a practical way (\textbf{Section \ref{subsec:RQ4}})?
\end{itemize}
Before that, we first introduce the experiment settings in \textbf{Section \ref{subsec:settings}}. 

\subsection{Settings}
\label{subsec:settings}
\subsubsection{Dataset} 
Table \ref{tab:dataset_info} lists the statistics of the 3 real-world trip
data on Beijing, Xi’an, and Chengdu. Beijing dataset \cite{10.1145/3183713.3183743} contain taxi trajectories; Xi'an and
Chengdu datasets contain trajectories by DiDi ride-sharing\footnote{https://gaia.didichuxing.com/}. Table \ref{tab:dataset_info} also provides the time span of each trip and the number of grid cells for different grid sizes.

\begin{table}[ht]
  \centering
  \caption{Dataset information}
  \label{tab:dataset_info}
  \small
  \setlength{\tabcolsep}{1pt} 
  \begin{tabular}{>{\raggedright\arraybackslash}p{3cm} p{2cm} p{2cm} p{2cm}} 
    \toprule
    Dataset & Beijing & Xi'an & Chengdu \\
    \midrule
    \midrule
    \#Trip & 3,100,845 & 2,419,072 & 3,887,769 \\
    \midrule
    Time span & 2009/3/2-2009/3/25 & 2016/10/1-2016/10/31 & 2016/10/1-2016/10/31 \\
    \midrule
    \#Grid Cells ($1,000m\times 1,000m$) & 250,120 & 14,001 & 30,028 \\
    \midrule
    \#Grid Cells ($2,000m\times 2,000m$) & 140,160 & 8,103 & 17,745 \\
    \bottomrule
  \end{tabular}
\end{table}

Furthermore, Points of Interest (POIs) data were obtained through APIs provided by Tencent map API\footnote{https://lbs.qq.com/service/webService} \DEL{\YH{which one did we use? Please add the name}},
utilizing the latitude and longitude of different grid cells. For more detailed POI categories descriptions, please refer to the Appendix.

\subsubsection{Baseline Models}
\begin{itemize}
    \item \textbf{Random Forest} (RF) \cite{pourebrahim2019trip} is a traditional machine learning method that consists of an ensemble of multiple decision tree models. Each tree is built on a subset of the data, and the final prediction is made by aggregating the predictions of all the individual trees.
  \item \textbf{Gravity Model} \cite{lenormand2016systematic} is a traditional
spatial interaction model which is inspired by Newton’s law
of Gravitation. The spatial features of one region work as
the mass and population flow between two regions follow
the power-law distance decay.   
    \item \textbf{Gradient Boosted Regression Trees}(GBRT) \cite{robinson2018machine}  is a machine learning method that builds an ensemble of decision trees sequentially, where each new tree is trained to correct the errors made by the previous ones.
    \item \textbf{GODDAG} \cite{rong2023goddag} is a deep learning method designed to generate origin-destination (OD) flow data for cities where such data is unavailable. It leverages graph neural networks (GNN) to model spatial dependencies and employs domain adversarial training to transfer knowledge from a source city with ample data to a target city with scarce data, thereby improving the model's ability to generalize across different urban environments.
\end{itemize}

\vspace{-0.1in}
\begin{table}[ht]
\caption{Baseline models}
\label{table:your_label}
\centering{
\small %
\setlength{\tabcolsep}{2pt}
\begin{tabular}{lll}
\hline
\textbf{Models} & \textbf{Techniques}\\ \hline
Random Forest \cite{pourebrahim2019trip} & Tree-based Model \\
Gravity Model \cite{lenormand2016systematic} & Physical Model  \\
GBRT \cite{robinson2018machine} & Tree-based Model   \\
GODDAG \cite{rong2023goddag}& Deep Learning\\ \hline
\end{tabular}
}
\end{table}
\vspace{-0.1in}

\subsubsection{Evaluation Metric}
We employ three widely-adopted metrics to assess the predictive efficacy of the models as  Rong\cite{rong2023goddag} did:
\begin{itemize}
  \item \textbf{Root mean square error(RMSE)} RMSE is a popular metric applied in regression problems. The formula for RMSE is given by equation \ref{RMSE}, where $f^X_{ij}$ are the observed values, $\hat{f}^X_{ij}$ are the predicted values, and $\left|\mathbf{F}^X\right|$ is the number of elements in the OD matrix $\mathbf{F}^X$, i.e., $\left|\mathbf{F}^X\right| = N^X \times N^X$.
  \begin{equation}
    \label{RMSE}
    RMSE = \sqrt{\frac{1}{\left|\mathbf{F}^X\right|} \sum_{i,j} \left(f^X_{ij} - \hat{f}^X_{ij}\right)^2}
    \end{equation}
  \item \textbf{Symmetric Mean Absolute Percentage Error (SMAPE)} SMAPE is used to measure the accuracy of forecasts. It is a variation of the Mean Absolute Percentage Error (MAPE) that addresses some of the limitations of MAPE, particularly its asymmetry and sensitivity to scale. The value range is in [0,2]. Its formula is given in equation \ref{SMAPE} with lower values indicating better forecasting accuracy.
    \begin{equation}
    \label{SMAPE}
    SMAPE = \frac{100\%}{\left|\mathbf{F}^X\right|} \sum_{i,j} \frac{\left|f^X_{ij} - \hat{f}^X_{ij}\right|}{\left(\left|f^X_{ij}\right| + \left|\hat{f}^X_{ij}\right|\right)/2}
    \end{equation}
  \item \textbf{Common Part of Commuters (CPC)} CPC measures the similarity between two data sets by comparing their common parts to the sum of their individual parts. The range is in [0,1], with a value of 1 indicating that the two commuting patterns being compared are identical and all commuters in one pattern are also present in the other pattern. Its formula is given in equation \ref{CPC}.
  \begin{equation}
    \label{CPC}
    CPC = \frac{2 \sum_{i,j} \min\left(f^X_{ij}, \hat{f}^X_{ij}\right)}{\sum_{i,j} \hat{f}^X_{ij} + \sum_{i,j} f^X_{ij}}
\end{equation}
\end{itemize}

\subsubsection{Experiment Settings}

The experiments were carried out on a system with 4 NVIDIA A100 GPUs, each with 40GB of memory. For training LLM based models, we used the Adam optimizer with a learning rate of 0.001, while Random Forest, Gravity Model and GBRT used the Adam optimizer with 0.0002 learning rate and weight decay of 0.001. The LLMs used is LLAMA2 7B. The experiments were conducted with a batch size of 100.

\begin{table*}[ht]
\centering
\caption{Comparison of models for different cell sizes}
\small
\setlength{\tabcolsep}{4pt} 
\begin{tabular*}{\textwidth}{l @{\extracolsep{\fill}}| c c c | c c c || c c c | c c c}
\specialrule{1.0pt}{0pt}{0pt}
\multicolumn{1}{c|}{} & \multicolumn{6}{c||}{Cell size = 1,000m$\times$1,000m} & \multicolumn{6}{c}{Cell size = 2,000m$\times$2,000m} \\ \cline{2-13}
\multicolumn{1}{c|}{} & \multicolumn{3}{c|}{Beijing $\rightarrow$ Chengdu} & \multicolumn{3}{c||}{Beijing $\rightarrow$ Xi'an} & \multicolumn{3}{c|}{Beijing $\rightarrow$ Chengdu} & \multicolumn{3}{c}{Beijing $\rightarrow$ Xi'an} \\ \cline{2-13}
Methods  & RMSE & SMAPE & CPC & RMSE & SMAPE & CPC & RMSE & SMAPE & CPC & RMSE & SMAPE & CPC \\ \hline
 \hline
Random Forest      & 25.32 & 2.00 & 0.00 & 8.72 & 2.00 & 0.00 & 304.85 & 2.00 & 0.00 & 444.25 & 2.00 & 0.00 \\ \hline
Gravity Model     & 13.08 & 2.00 & 0.00 & 10.56 & 2.00 & 0.00 & 50.51 & 2.00 & 0.00 & 67.8 & 2.00 & 0.00 \\ \hline
GBRT    & 29.12 & 2.00 & 0.00 & 9.07 & 2.00 & 0.00 & 268.09 & 2.00 & 0.00 & 446.26 & 2.00 & 0.00 \\ \hline
GODDAG  & 4.29 & 2.00 & 0.00 & 3.94 & 2.00 & 0.00 & 26.97 & 2.00 & 0.00 & 28.87 & 2.00 & 0.00 \\ \hline
\hline
LLM-COD & \textbf{2.29} & \textbf{0.00} & \textbf{0.57} & \textbf{2.12} & \textbf{0.00} & \textbf{0.57} & \textbf{24.25} & \textbf{0.00} & \textbf{0.63} & \textbf{28.49} & \textbf{0.00} & \textbf{0.42} \\ \hline
\end{tabular*}
\label{tab:comparison_models}
\end{table*}

\subsection{Cross City OD Flow Prediction (RQ1)}
\label{subsec:RQ1}
To evaluate the performance of our novel model in cross-city prediction tasks, we trained our model on Beijing datasets and performed prediction on on Chengdu and Xi'an datasets, which were not seen during the training phase. The results are summarized in Table \ref{tab:comparison_models}. We have the following observations:
\begin{itemize}
\item[(1)]\textbf{Consistent Superiority in Both Cities} Our model consistently outperforms the comparison method in both Chengdu and Xi'an. Notably, our model exhibits significant advantages in terms of RMSE, SMAPE, and CPC metrics, indicating its proficiency in maintaining the distribution of origin-destination (OD) flow in cross-city prediction scenarios. The performance in Chengdu is slightly better than Xi'an. This is attributed to the similarities between Chengdu and Beijing, including urban grid structure and population density. One of the reason that the traditional models and GODDAG underperform may be that they cannot effectively utilize the origin set of the target city because of the structure design the models, resulting in low high SMAPE and low CPC.
\item[(2)] \textbf{LLM-CCD Excels in High Precision OD Flow Prediction} LLM-CCD reduces the RMSE of  the current best model GODDAG by ~46\% for grid size of 1,000 m $\times$ 1,000m. The ability to predict origin-destination trip flow for high spatial resolution is critical to optimizing transportation systems.
\end{itemize}
\vspace{-0.0in}
\subsection{Performance on Design Choices (RQ2)}
\label{subsec:RQ2}

To assess the performance of our proposed LLM-COD framework, we conducted experiments focusing on different design choices. Our evaluations utilized the Beijing to Xi'an dataset. The experiments compare the performance of our LLM-COD model under three different settings: the impact of city indicators ``\emph{[city]}'' in the instruction format (as described in Table \ref{table:instruction}), the use of a Single POI strategy, and the impact of distance cost. 
\begin{itemize}
\item[(1)] LLM-COD with Full Features: This includes all city indicators and multiple POI types for each grid cell. 
\item[(2)] LLM-COD without City Indicators: This version omits city-specific indicators in the instruction format to evaluate their impact on performance. Specifically, the instruction format follows: ``<Instruction> \emph{Given the starting place and time of a taxi trajectory, predict the most likely destination and how far it is from the starting point.}''
\item[(3)] LLM-COD with Single POI Strategy: Each grid cell is represented by at most one POI type. If the cell originally has multiple POIs, the policy to choose is based on social check-in scenarios, prioritizing popular locations where users frequently check in.
\item[(4)] LLM-COD with diffrent $\alpha$ (the weight assigned to the travel cost $\hat{c}^X_{l}$ in the loss function (Equ. (\ref{eq:crossentropy}))).
\end{itemize}

As shown in Table \ref{tab:city_comparison}, the LLM-COD model with full features achieves an RMSE of 24.25, a SMAPE of 0.00, and a CPC of 0.42 when predicting OD flows from Beijing to Xi'an. When the city indicators are removed, the RMSE increases to 27.96, and the CPC slightly decreases to 0.40, indicating that city-specific features significantly enhance model performance. Furthermore, employing the Single POI strategy results in a higher RMSE of 31.28 and a CPC of 0.40, suggesting that representing each grid cell with multiple POIs provides a richer and more informative feature set, leading to better performance.

\begin{figure}[t]
\centering
\hspace{0.00in}
\begin{minipage}{0.49\textwidth}
  \subfigure{
\includegraphics[width=1.00\textwidth]{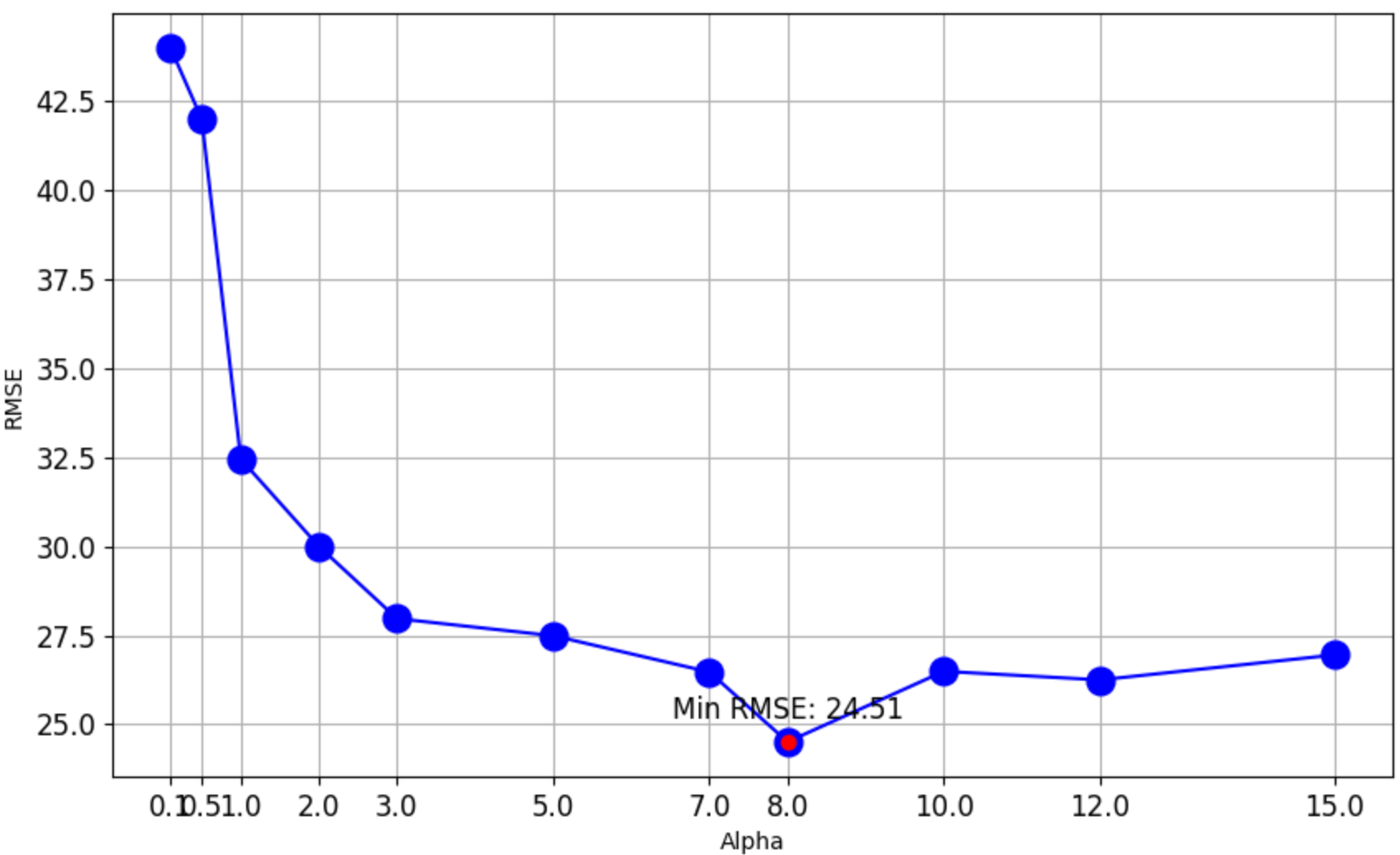}}
\vspace{-0.25in}
\end{minipage}
\caption{Effects of Alpha}
\label{fig:Alpha}
\vspace{-0.10in}
\end{figure}

As shown in Figure \ref{fig:Alpha}, with alpha increasing, the average RMSE first decreases then increases. This trend occurs because taxi drivers often prefer destinations with shorter travel distances. However, when the weight for distance becomes too large, the model tends to overemphasize proximity, consistently choosing the nearest destinations. This approach neglects destinations that are slightly further away but are preferred for their higher quality or better reviews. This behavior is particularly evident in categories such as restaurants and hotels, where quality and reputation often outweigh mere proximity.

\begin{table}[ht]
\centering
\caption{Performance in Different Model Parameters}
\small
\setlength{\tabcolsep}{1pt}
\begin{tabular}{ l|c c c||c c c }
\specialrule{1.0pt}{0pt}{0pt}
\multicolumn{1}{ c|}{} & \multicolumn{3}{c||}{Beijing $\rightarrow$ Chengdu} & \multicolumn{3}{c }{Beijing $\rightarrow$ Xi'an} \\ \cline{2-7}
Methods  & RMSE & SMAPE & CPC & RMSE & SMAPE & CPC \\ \hline
 \hline
LLM-COD    &\textbf{28.49}  & \textbf{0.00}  & \textbf{0.63}  & \textbf{24.25}  & \textbf{0.00}  & \textbf{0.42}\\  \hline
LLM-COD (without city indicators) &32.96   &0.00  & 0.57  & 27.96  &0.00  &0.40    \\  \hline
LLM-COD (Single POI)    &32.95  & 0.00  & 0.57  &31.28   & 0.00  &0.40  \\ \hline
\end{tabular}
\label{tab:city_comparison}
\end{table}

\vspace{-0.0in}
\subsection{Robustness Analysis (RQ3)}
\label{subsec:RQ3}

\begin{figure*}[htbp]
    \centering
    \subfigure{
        \includegraphics[width=0.48\textwidth]{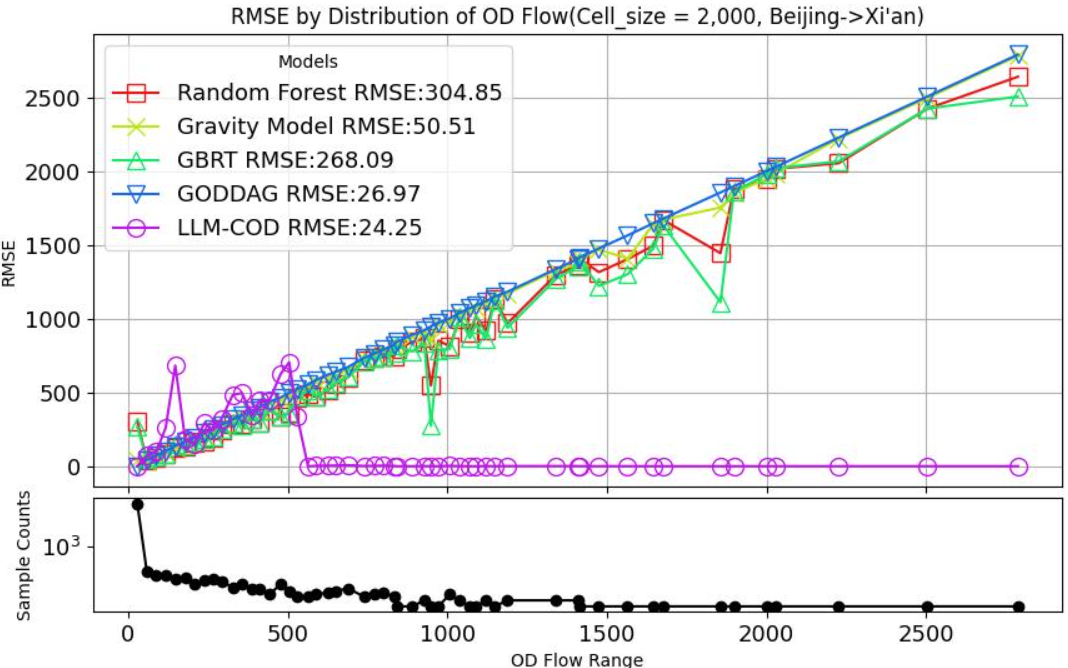}
        \label{subfig:OD_2000_BJ-XA}}
    \subfigure{
        \includegraphics[width=0.48\textwidth]{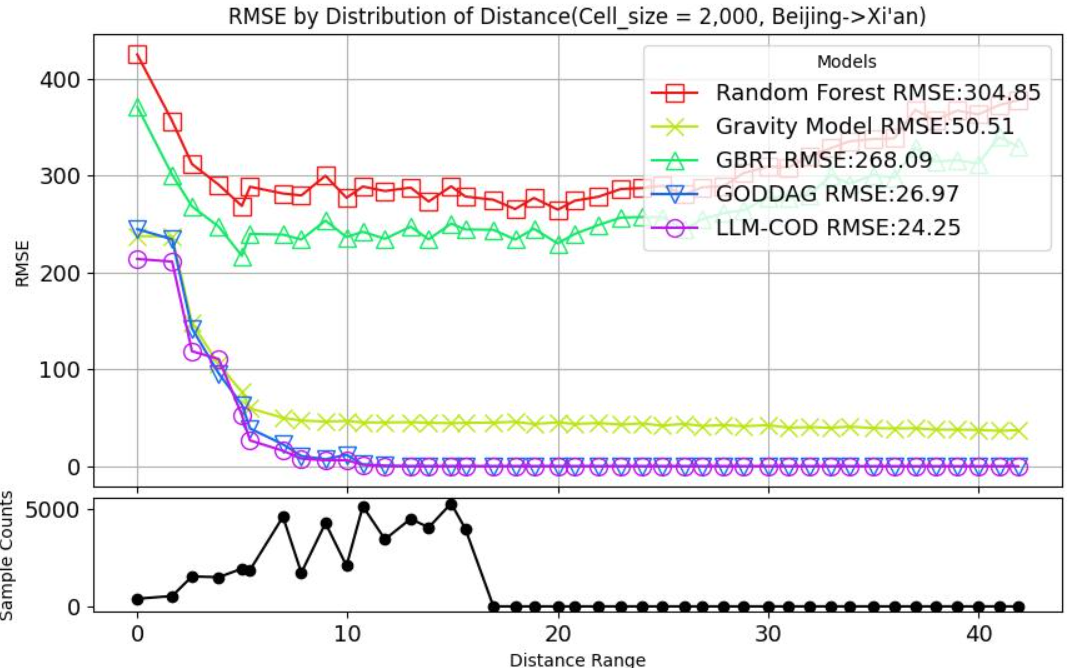}
        \label{subfig:distance_2000_BJ-XA}
    }
    \subfigure{
        \includegraphics[width=0.48\textwidth]{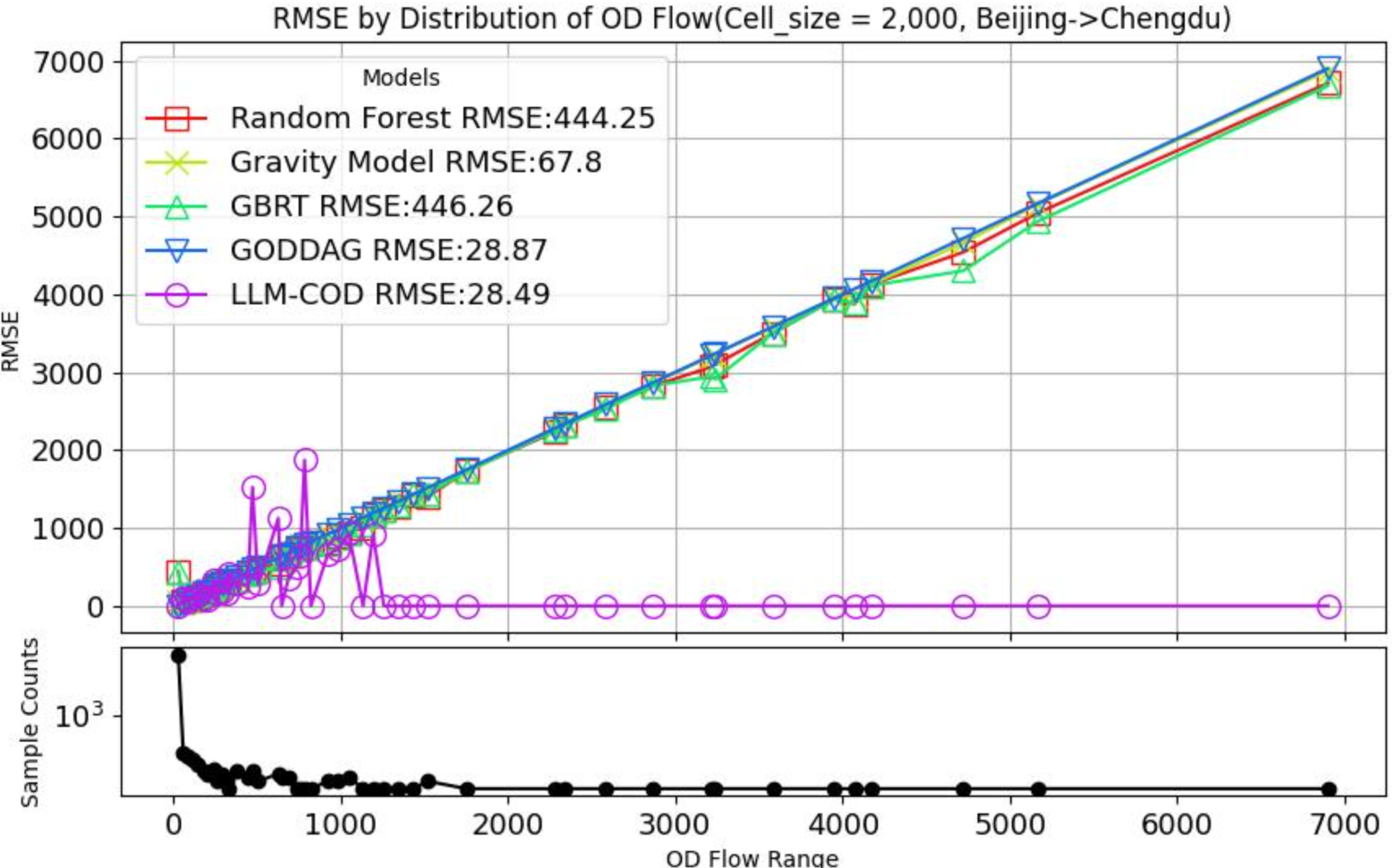}
        \label{subfig:OD_2000_BJ-CD}}
    \subfigure{
        \includegraphics[width=0.48\textwidth]{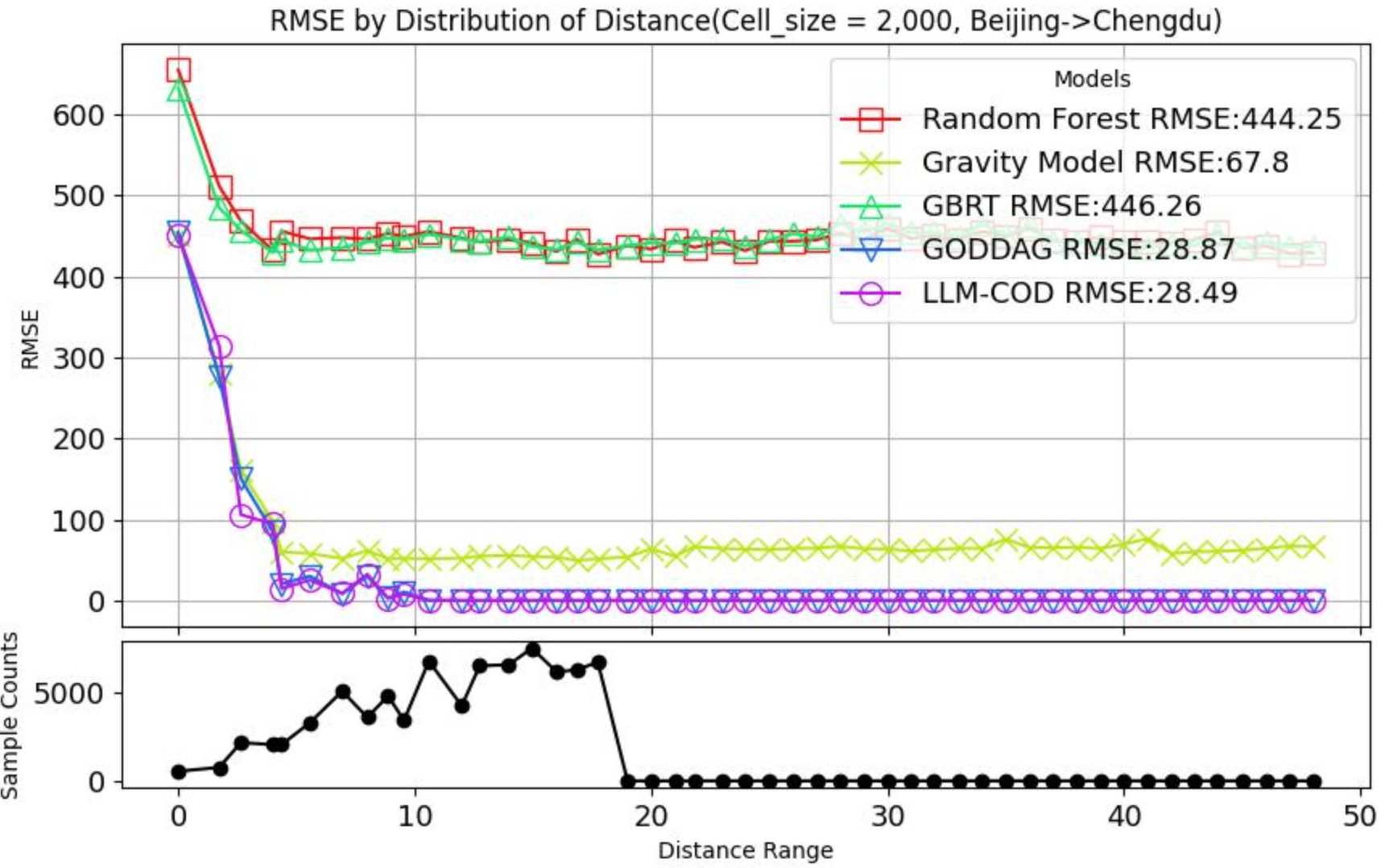}
        \label{subfig:distance_2000_BJ-CD}}
        \vspace{-0.1in}
    \caption{Robustness Analysis ($2,000m\times 2,000m$)} 
    \label{fig:robustness_RQ3_2}
    \vspace{-0.0in}
\end{figure*}

To evaluate the robustness of our proposed LLM-COD model, we conducted a detailed analysis focusing on the distribution of OD flow and travel distance across the entire test dataset. We specifically examined the RMSE performance as a function of OD flow and distance, across four experimental setups: two different cell sizes (1,000m and 2,000m)  and two city pairs (Beijing to Chengdu and Beijing to Xi'an). Figures \ref{fig:robustness_RQ3_1} and \ref{fig:robustness_RQ3_2} present the results of this robustness analysis for cell sizes 1,000m and 2,000m respectively. The key observations from these experiments are as follows:

\begin{figure*}[htbp]
    \centering
    \subfigure{
        \includegraphics[width=0.48\textwidth]{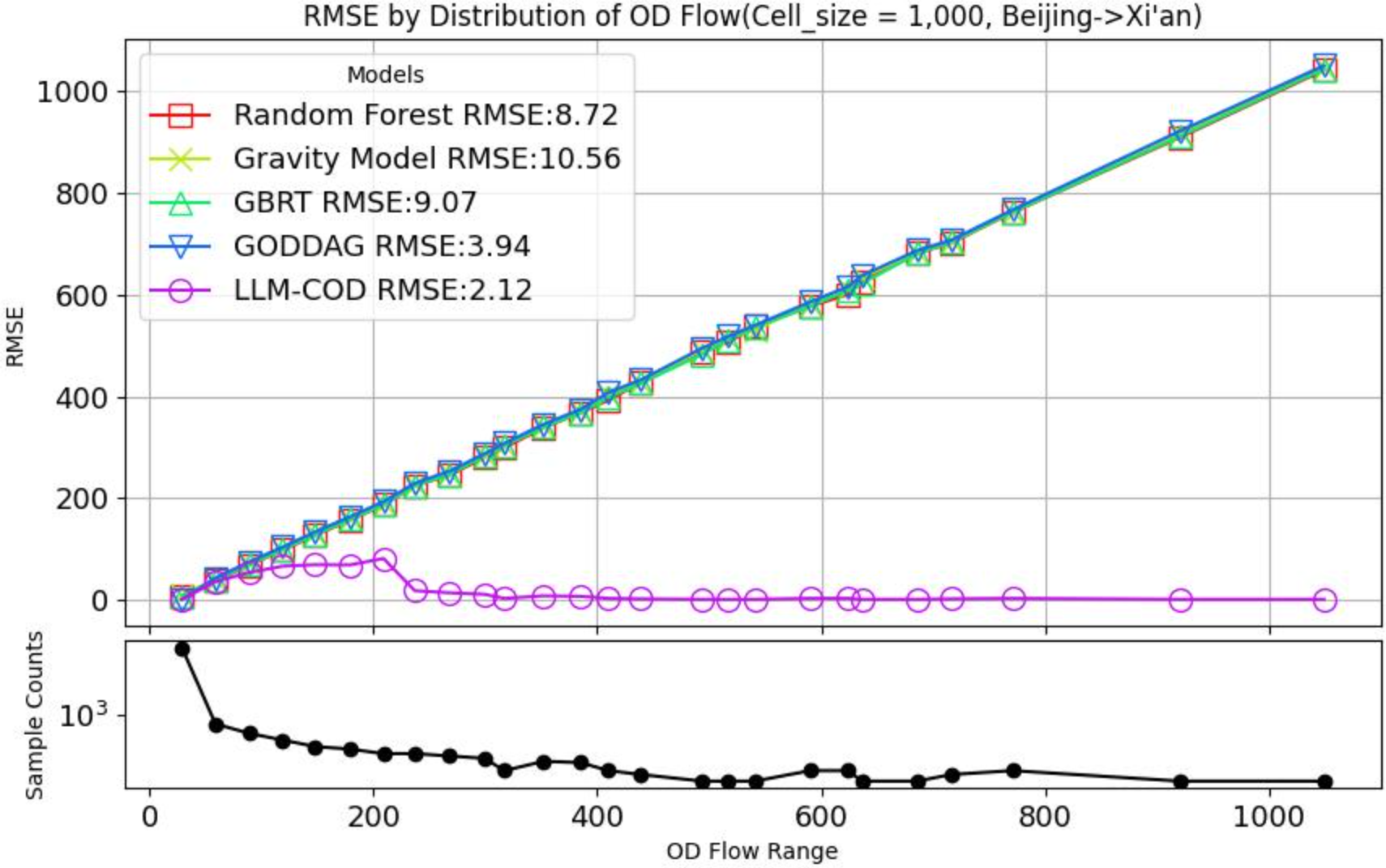}
        \label{subfig:OD_1000_BJ-XA}}
    \subfigure{
        \includegraphics[width=0.48\textwidth]{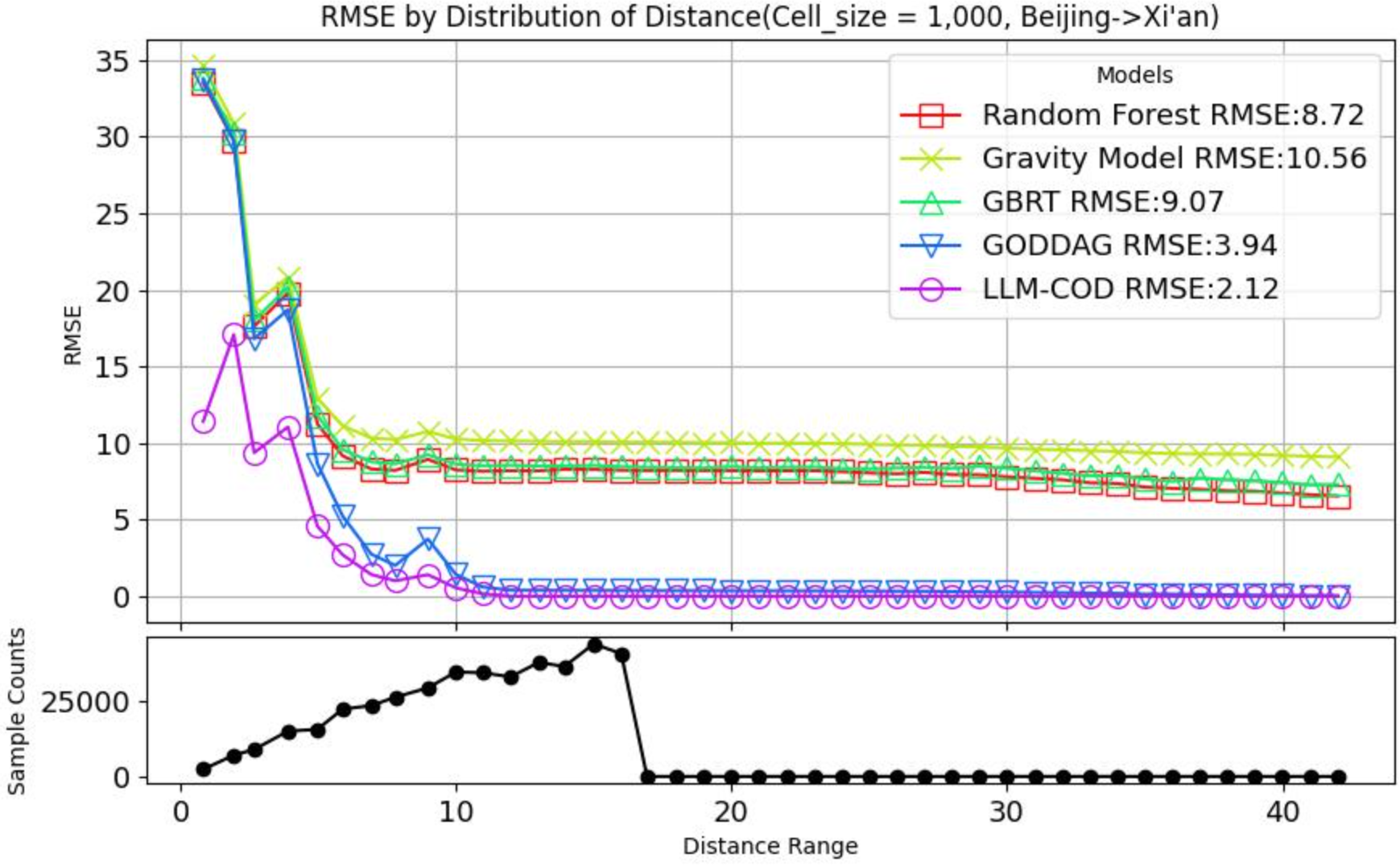}
        \label{subfig:distance_1000_BJ-XA}
    }
    \subfigure{
        \includegraphics[width=0.48\textwidth]{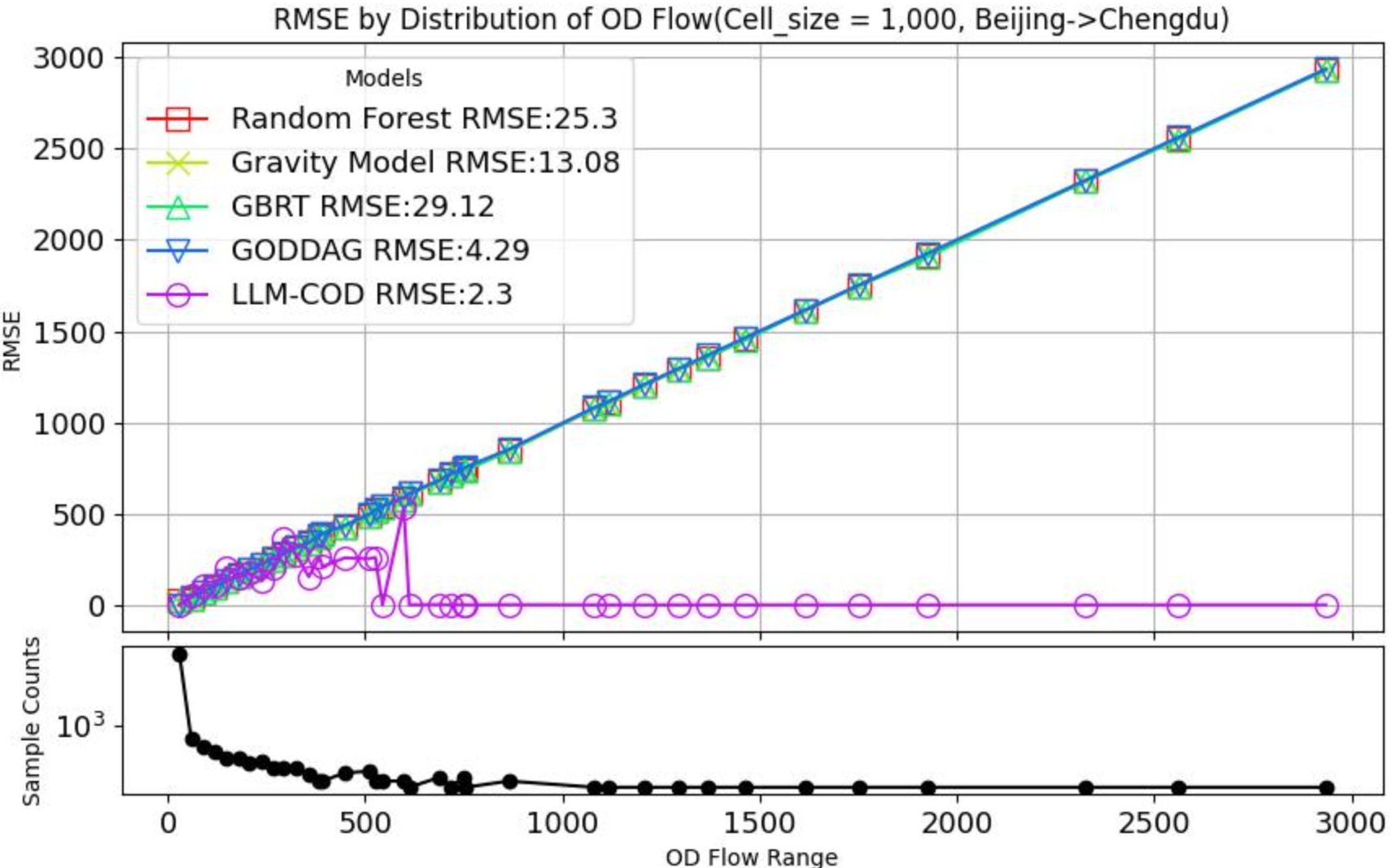}
        \label{subfig:OD_1000_BJ-CD}}
    \subfigure{
        \includegraphics[width=0.48\textwidth]{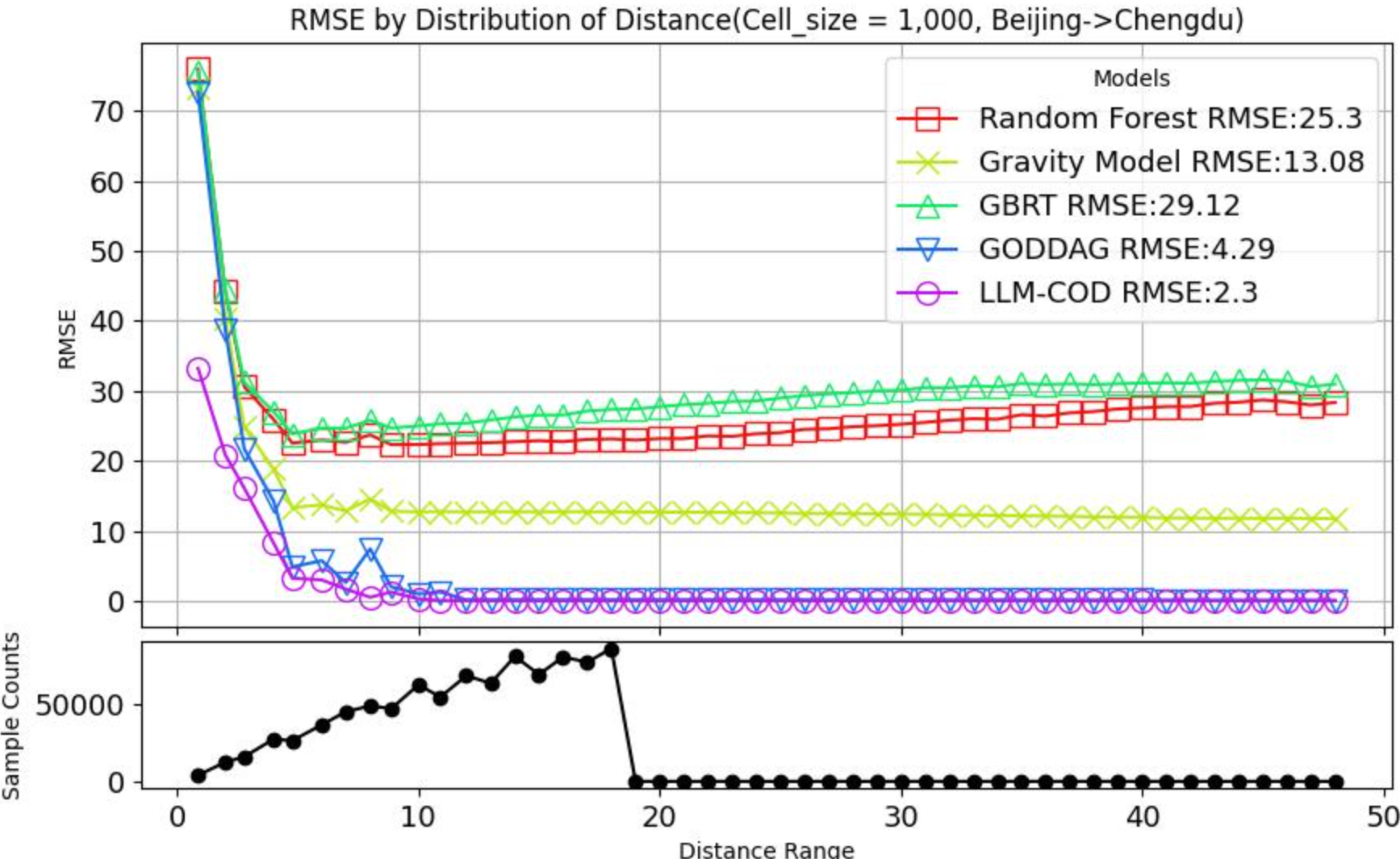}
        \label{subfig:distance_1000_BJ-CD}}
        \vspace{-0.1in}
    \caption{Robustness Analysis ($1,000m\times 1,000m$)} 
    \label{fig:robustness_RQ3_1}
    \vspace{-0.1in}
\end{figure*}

\begin{itemize}
\item[(1)] The LLM-COD model consistently achieves a RMSE of zero as the OD flow or distance increases. This indicates that our model effectively captures and predicts high-volume and long-distance flows, maintaining accuracy even as these metrics increase. This trend is observed across all eight experiment settings, demonstrating the robustness of our model.
\item[(2)] On the distribution of OD flow, the GODDAG model, along with other baseline models, exhibit large errors and considerable deviation from the actual distribution. These models struggle with both high and low OD flows, almost get the identical error as the flow value. 
\item[(3)] On the distribution of distance, GODDAG, represented by blue markers, achieves a RMSE of zero as the distance value increases. But it shows slightly larger errors in the mid-range distances compared to LLM-COD.
\item[(4)] The validity of the LLM-COD model can be attributed to its ability to preserve the underlying distribution of OD flows as found in the ground truth. Unlike other models, LLM-COD leverages comprehensive city indicators and POI data, enabling it to generalize well across different cities, effectively transferring learned mobility patterns from the source city to the target city.
\end{itemize}
\vspace{-0.0in}
\subsection{Visualization (RQ4)}

\begin{figure*}[htbp]
    \centering
    \subfigure[Ground Truth]{
        \includegraphics[width=0.32\textwidth, height = 5.3cm]{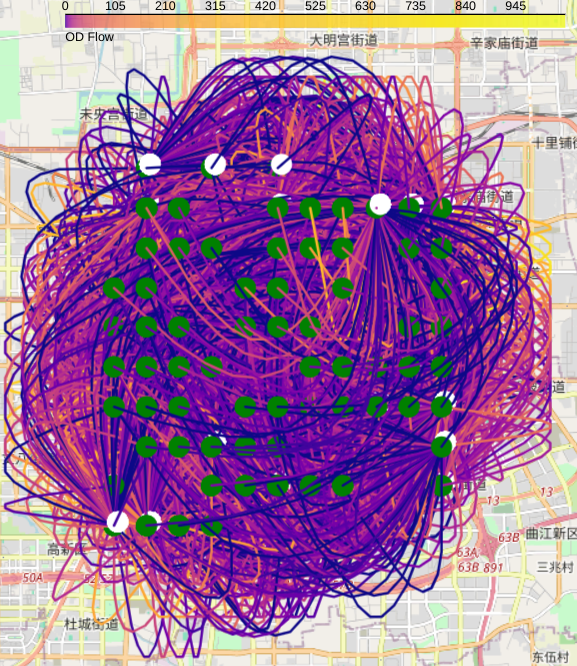}
        \label{subfig:ground_truth}}
    \subfigure[Random Forest]{
        \includegraphics[width=0.32\textwidth, height = 5.3cm]{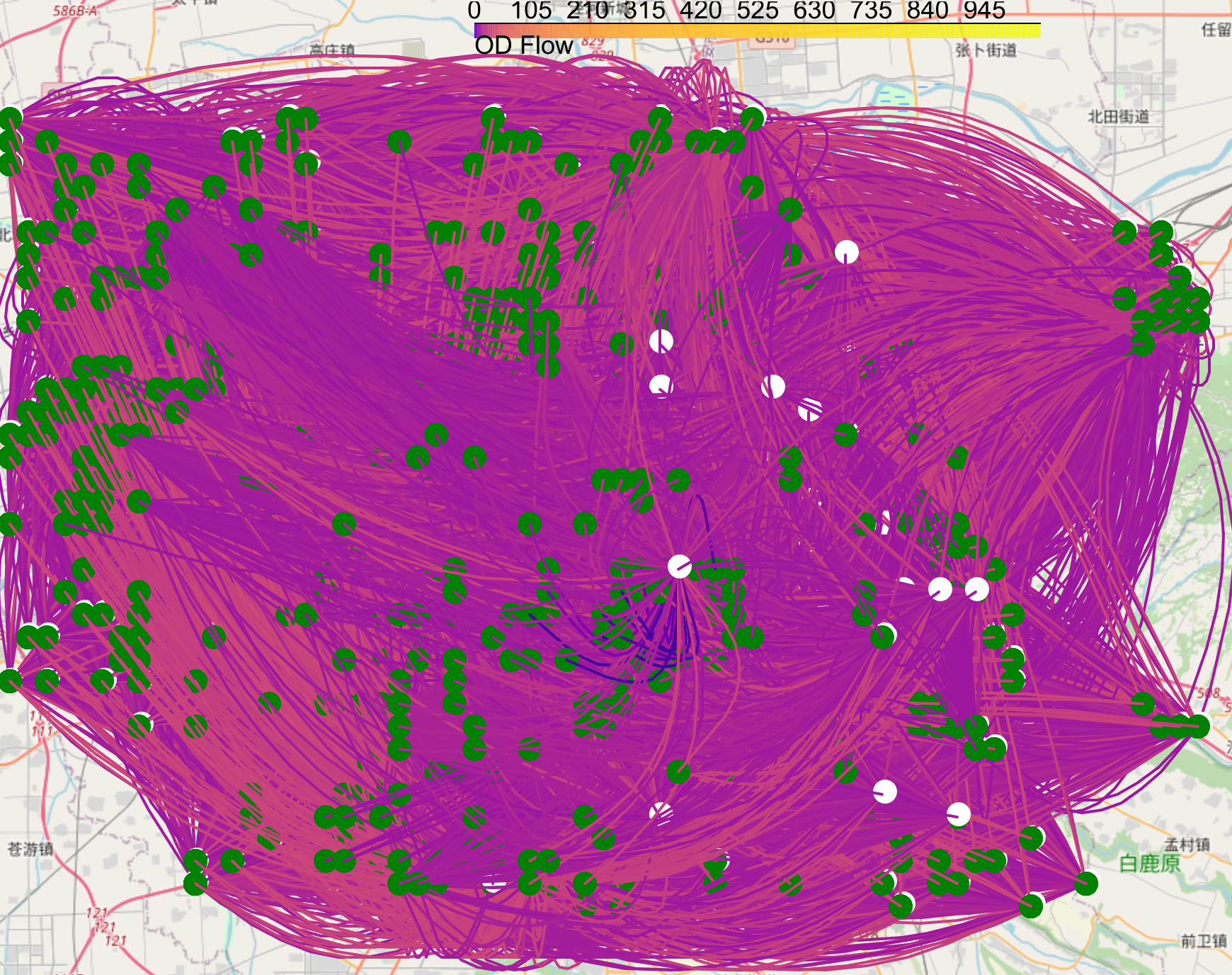}
        \label{subfig:randomforest}}
    \subfigure[Gravity Model]{
        \includegraphics[width=0.32\textwidth, height = 5.3cm]{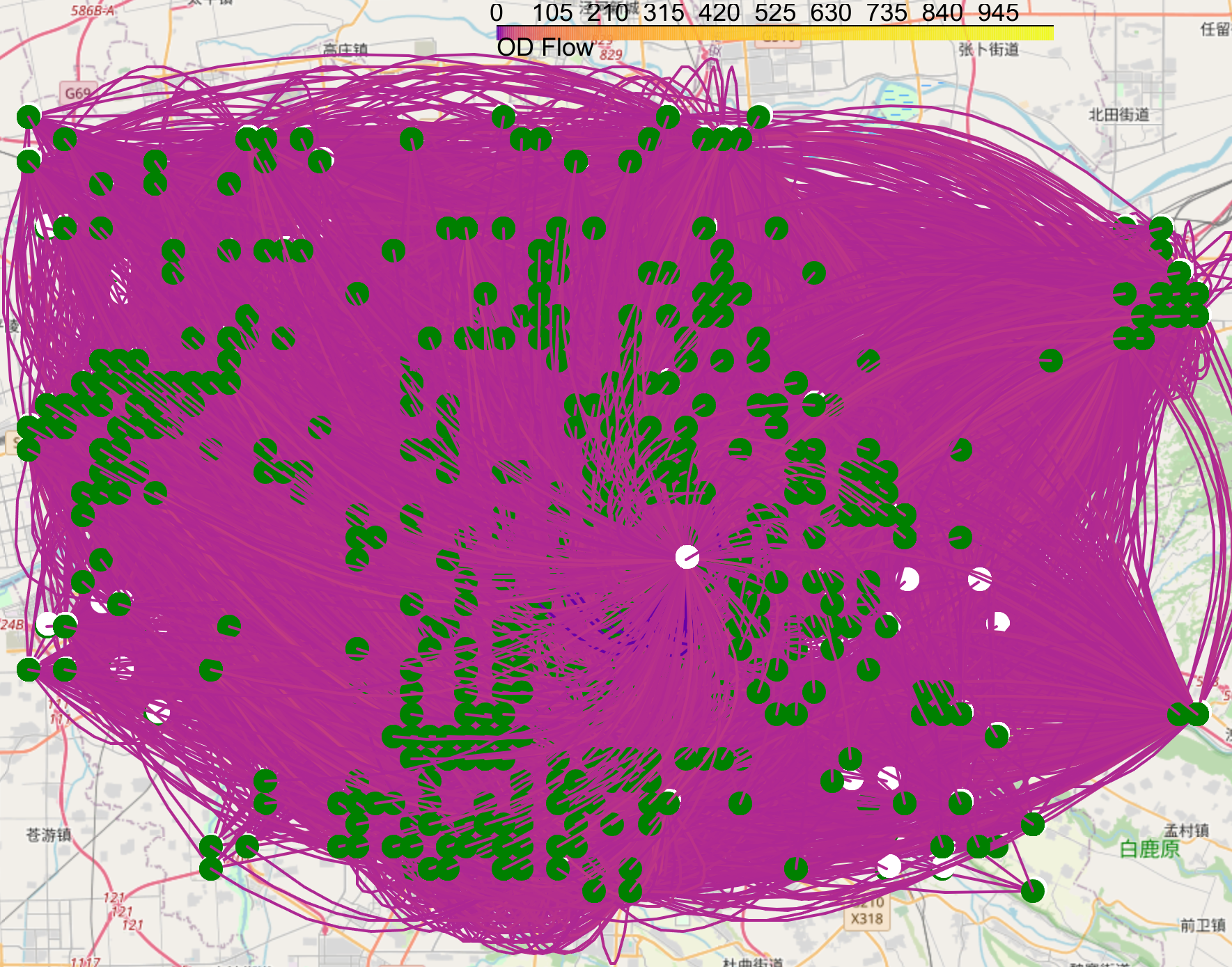}
        \label{subfig:gravity}}
    \subfigure[GBRT]{
        \includegraphics[width=0.32\textwidth, height = 5.3cm]{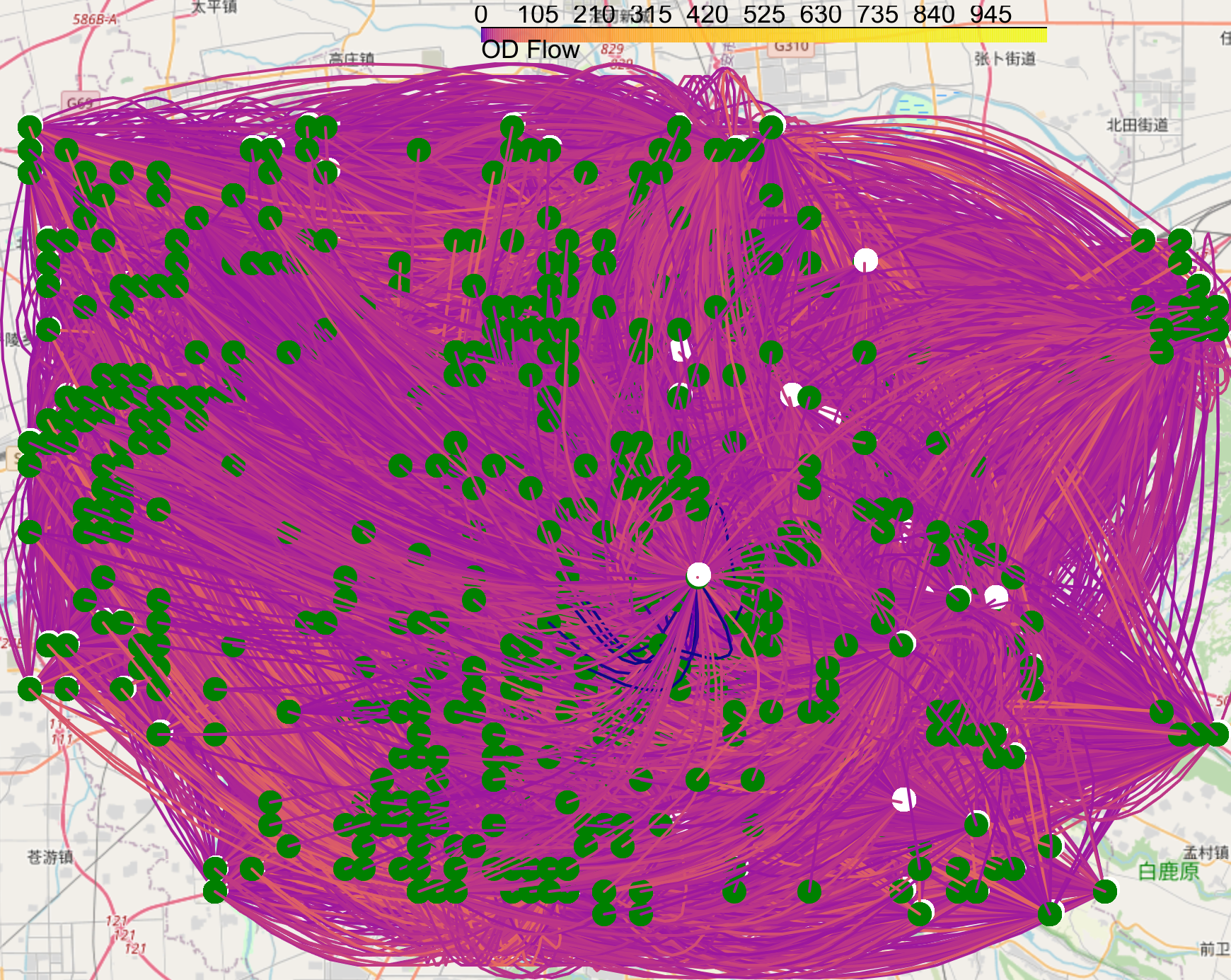}
        \label{subfig:GBRT}}
    \subfigure[GODDAG]{
        \includegraphics[width=0.32\textwidth, height = 5.3cm]{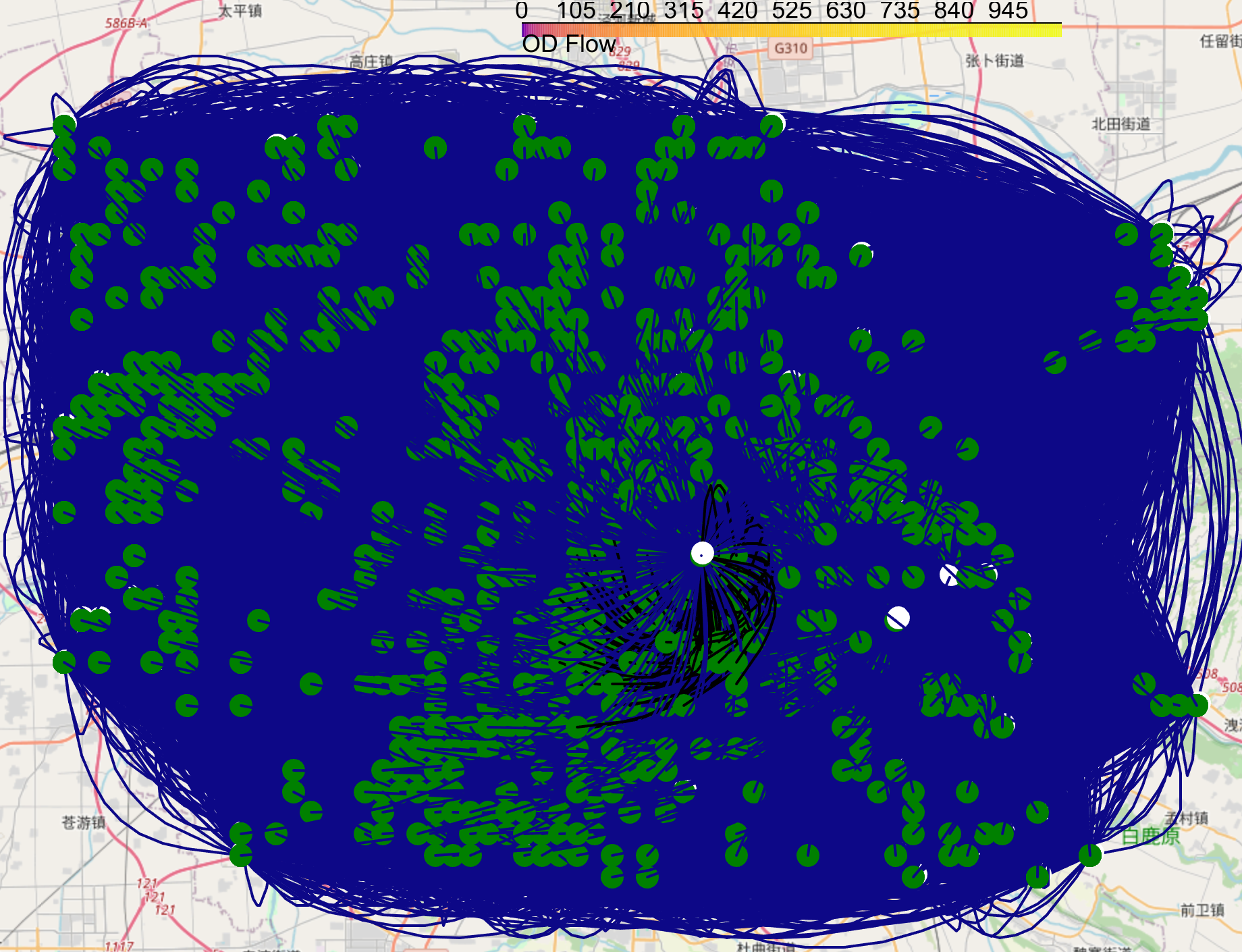}
        \label{subfig:GOD}}
    \subfigure[LLM-COD]{
        \includegraphics[width=0.32\textwidth, height = 5.3cm]{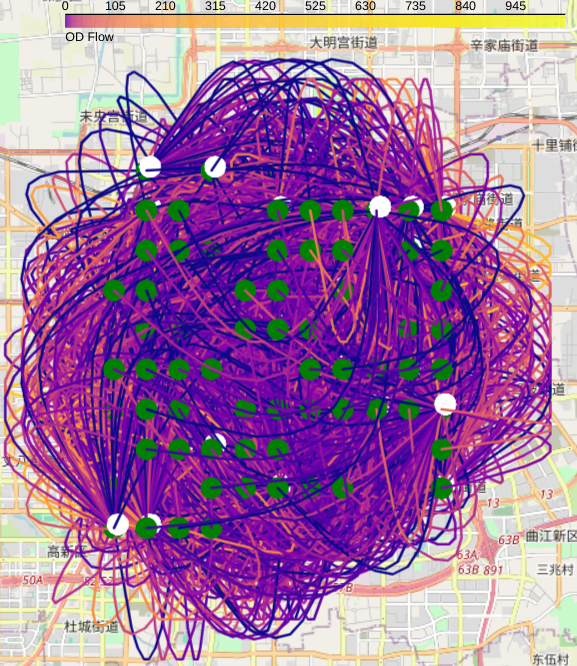}
        \label{subfig:ours}
    }
    \vspace{-0.1in}
    \caption{Visualization of the generated OD flow of ground truth and different models under cell size = 2,000m (Beijing -> Xi'an), where green marker stands for origin and white marker stands for destination. The diagrams from left to right are from (a) Ground truth, (b) Random Forest, (c) Gravity Model, (d) GBRT, (e) GODDAG, (f) LLM-COD}
    \label{fig:visualization_RQ4}
    \vspace{-0.0in}
\end{figure*}

To further demonstrate the effectiveness of our proposed OD generation framework and provide an intuitive comparison, we visualize the generated OD flows from our model and several baselines against the ground truth. For this comparison, we select Xi'an as the target city and Beijing as the source city. The cell size is set to 2,000m. As illustrated in Figure \ref{fig:visualization_RQ4}, OD flows are depicted using an arc diagram, where each arc represents a flow between two locations. Green marker stands for origin and white marker stands for destination. The brightness of the arcs indicates the volume of OD flows, with blue indicates small volumes, red indicates medium volumes, and yellow indicates the largest volumes of OD flows.

In Figure \ref{fig:visualization_RQ4}, the ground truth shows that most taxi trips starting from the surrounding areas to the city center. Our LLM-COD model (Figure \ref{fig:visualization_RQ4}f) closely replicates this spatial distribution, aligning well with the ground truth.

Conversely, the GODDAG model (Figure \ref{fig:visualization_RQ4}c), mainly shows blue arcs, indicating smaller volumes. However, the distribution of OD flows is incorrect, with the predicted OD flows forming an elliptical shape, suggesting that people have longer commuting ranges in the northwest-southeast direction, which is not true for Xi'an. This leads to a lower RMSE for GODDAG when the distance becomes larger, but it fails to accurately represent the actual commuting patterns. Similarly, the Random Forest (Figure \ref{fig:visualization_RQ4}b), gravity model (Figure \ref{fig:visualization_RQ4}c) and GBRT model (Figure \ref{fig:visualization_RQ4}d), also display elliptical distributions of OD flows, which is inconsistent with the actual commuting patterns in Xi'an. These models show predominantly red arcs, indicating larger volumes of OD flows but also higher errors. Additionally, it is difficult to discern the true center of the city from the distributions generated by these baseline models.

This visual comparison underscores the robustness and accuracy of our LLM-COD model in replicating the true distribution of OD flows.
\label{subsec:RQ4}
\vspace{-0.0in}
\section{Related Work}

In the domain of OD flow prediction, various approaches have been developed to forecast OD flows using accessible urban data. This field has evolved from \emph{traditional models} (Section \ref{subsec:relatedtradition}) to advanced \emph{learning-based models} (Section \ref{subsec:relatedlearning}), adapting to the complexities of urban mobility patterns.

\subsection{Traditional Models}
\label{subsec:relatedtradition}
Initial research predominantly employed models such as the Gravity Model \cite{lenormand2016systematic} and the Radiation Model \cite{ren2014predicting}. The Gravity Model posits that the OD flow is proportional to the population sizes of the origin and the destination and inversely proportional to the square of their distance. Recent extensions of this model have incorporated novel data sources such as social media check-ins from X \cite{pourebrahim2019trip}, enhancing the model's adaptability to modern urban dynamics. The Radiation Model introduces a probabilistic approach \cite{ren2014predicting}, considering the attractiveness of intermediate regions, thus providing insights into spatial dynamics. This model has also been refined to include socioeconomic variables, such as income and education, improving its predictive performance \cite{lenormand2016systematic}. These developments signify a methodological evolution, adapting traditional models to leverage the increasing availability of rich urban data.

Traditional models might fall short in prediction performance, yet they provide essential insights and intuition for building predictive models. Learning-based models, noted for their flexibility and complex structures, deliver commendable performance, especially in capturing implicit data characteristics. 
\vspace{-0.0in}
\subsection{Learning-Based Models} \label{subsec:relatedlearning}
The advent of machine learning ushered a shift towards data-driven methodologies in OD prediction. Techniques such as Gradient Boosted Regression Trees (GBRT) \cite{robinson2018machine} and Random Forest \cite{pourebrahim2019trip} have demonstrated superior capability in capturing complex interactions within mobility data. Further advancements in deep learning have led to the incorporation of graph-based methodologies, acknowledging the networked nature of urban spaces. Graph Machine Learning Ensemble (GMEL) \cite{liu2020learning} and Spatial Graph Attention Networks (Spatial GAT) \cite{cai2022spatial} leverage Graph Attention Networks (GAT) \cite{velivckovic2017graph} and Graph Convolutional Networks (GCN) \cite{kipf2016semi} to model the intricate spatial relationships essential for accurate OD flow prediction. These approaches significantly enhance the ability to interpret the influence of both direct and neighboring regions on mobility patterns.\\



Despite these advancements,  these works all target predicting OD flows in a single city, while achieving OD prediction across cities remains an unresolved issue. 
Notably, OD prediction models tailored for a single city \cite{9657493, 9805695} often depend on unique urban characteristics, such as transportation modes and urban layouts, making it challenging to apply the learned features from one city to another. The work most related to our works is proposed by Rong et al. \cite{rong2023goddag}, which employes unsupervised transfer learning to enhance model transferability. This approach, named GODDAG, involves unsupervised domain adaptation, which trains models within a data-rich source domain and then adapts them to an unlabeled target domain. The central concept is to align features from both the source and target domains into a shared feature space, thereby minimizing the domain discrepancies. Consequently, this method enables the application of OD prediction models, initially trained in one urban environment, to be effectively used in another without the need for labeled data in the target domain. 

Leveraging the exceptional pattern recognition and inference capabilities of LLMs, our research introduces a new approach for predicting OD flows. By combining LLMs with traditional transfer learning techniques, we enhance model generalization across diverse urban environments. This novel method is particularly beneficial for rapidly evolving cities with limited data, providing a new solution to contemporary urban planning challenges.
\vspace{-0.0in}
\section{Conclusions}

We introduced LLM-COD which is a new method for cross-city OD flow prediction. By fine-tuning an LLM with a new loss function, our method can understand spatial and functional relationships within urban spaces and capture interactions between individuals and various POIs. Experimental results based on data from three major cities—Beijing, Chengdu, and Xi'an—demonstrate that LLM-COD outperforms state-of-the-art learning-based methods in cross-city OD flow prediction, especially for high-volume and long-distance flows. 

We envision several promising research directions to explore further. First, we are committed to enhancing the capabilities of our LLM-COD model by applying to more urban dataset. This will include integrating diverse urban datasets from different cities or even different coutries. Second, Understanding and interpreting the decision-making process of LLM-COD is another crucial area of focus. Although our model achieves high performance, it is essential to provide users with clear and understandable explanations of its predictions. In summary, our future endeavors will focus on expanding the data diversity and contextual understanding of LLM-COD, enhancing its interpretability.

\vspace{-0.0in}
\section{Acknowledgements}
The research was partially sponsored by the Army Research Laboratory and was accomplished under Cooperative Agreement Number W911NF-23-2-0014. The views and conclusions contained in this document are those of the authors and should not be interpreted as representing the official policies, either expressed or implied, of the Army Research Laboratory or the U.S. Government. The U.S. Government is authorized to reproduce and distribute reprints for Government purposes, not withstanding any copyright notation herein.
This research was partially supported by U.S. NSF grants CNS-2136948 and CNS-2313866.

\newpage 

\bibliographystyle{ACM-Reference-Format}
\bibliography{reference}

\newpage 
\appendix
\section{Experimental Details Description}
\subsection{POI category Details}

We collected POI data using tencent map API. The POI data we use contains 24 categories, as table \ref{tab:poi_frequency} shows.

\begin{table}[ht]
\centering
\caption{POI Types and Their Frequencies}
\label{tab:poi_frequency}
{\small %
\setlength{\tabcolsep}{2pt} %
\begin{tabular}{l@{\hspace{10pt}}l}
\hline
\textbf{POI Type} & \textbf{Frequency}\\ \hline
Residential Area & 105,183\\
Food \& Cuisine & 66,158\\
Commercial Building & 46,772\\
Infrastructure & 38,313\\
Tourist Attraction & 34,648\\
Organization & 34,302\\
Education \& School & 29,849\\
Hotel & 28,866\\
Shopping & 22,972\\
Healthcare & 18,315\\
Company \& Enterprise & 16,377\\
Industrial Park & 13,789\\
Automobile & 11,576\\
Real Estate Community Affiliated & 11,015\\
Sports \& Fitness & 9,715\\
Entertainment \& Leisure & 9,485\\
Cultural Venue & 8,420\\
Life Services & 5,536\\
Place Name \& Address & 4,226\\
Banking \& Finance & 2,559\\
Indoor \& Affiliated Facilities & 574\\
Other Real Estate Community & 2\\
Others & 1\\
\hline
\end{tabular}
} %
\end{table}

\section{LLM backend comparison}

When we started the project, Google hadn't released Gemma. On April 16th, Google made Gemma open source. As Gemma is more powerful than LLaMA2, we conducted a comparison between Gemma-7B and the LLaMA2 model we initially used. The table below presents the results of this comparison.

\begin{table}[h!]
\centering
\caption{Performance in Different LLM backend}
\label{tab:backend}
\small
\setlength{\tabcolsep}{1pt}
\begin{tabular}{ l|c c c||c c c }
\specialrule{1.0pt}{0pt}{0pt}
\multicolumn{1}{ c|}{} & \multicolumn{3}{c||}{Beijing $\rightarrow$ Chengdu} & \multicolumn{3}{c }{Beijing $\rightarrow$ Xi'an} \\ \cline{2-7}
Methods  & RMSE & SMAPE & CPC & RMSE & SMAPE & CPC \\ \hline
 \hline
LLM-COD    &28.49  & 0.00  & 0.63  & 24.25  & 0.00  & 0.42\\  \hline
LLM-COD(Gemma)    &\textbf{21.07}  & \textbf{0.00}  & \textbf{0.66}  & \textbf{19.18}  & \textbf{0.00}  & \textbf{0.48}\\  \hline
\end{tabular}
\end{table}

As shown in the Table \ref{tab:backend}, the LLM-COD model using Google Gemma significantly outperforms the one using LLaMA2. Specifically, the RMSE values for the OD flow predictions from Beijing to Chengdu and Xi'an are lower with Gemma, indicating improved accuracy. Similarly, the CPC values are higher, suggesting better alignment with the actual OD flow distribution. The experiments further demonstrate that our framework and new loss function are effective in cross-city OD flow prediction, showing increased performance with a better base LLM model and outperforming state-of-the-arts models further.

\end{CJK}

\end{document}